\documentclass[10pt,twocolumn,letterpaper]{article}

\usepackage[pagenumbers]{cvpr}
   
\PassOptionsToPackage{numbers, sort&compress}{natbib}

\usepackage[utf8]{inputenc} % allow utf-8 input
\usepackage[T1]{fontenc}    % use 8-bit T1 fonts
\usepackage{url}            % simple URL typesetting
\usepackage{booktabs}       % professional-quality tables
\usepackage{amsfonts}       % blackboard math symbols
\usepackage{nicefrac}       % compact symbols for 1/2, etc.
\usepackage{microtype}      % microtypography
\usepackage{xcolor}         % colors
\usepackage{capt-of}% or \usepackage{caption}
\usepackage{amsmath}
\usepackage{graphicx}
\usepackage{soul}
\usepackage{multirow}
\usepackage{pifont}
\usepackage{tabularx}
\usepackage{xcolor}
\usepackage{placeins}
\usepackage{xurl}
\usepackage{float}

\definecolor{cvprblue}{rgb}{0.21,0.49,0.74}
\usepackage[pagebackref,breaklinks,colorlinks,citecolor=cvprblue]{hyperref}

\newcommand{\refFig}[1]{Fig.~\ref{fig:#1}}
\newcommand{\refSec}[1]{Sec.~\ref{sec:#1}}
\newcommand{\refTbl}[1]{Tbl.~\ref{tbl:#1}}
\newcommand{\refEq}[1]{Eq.~\ref{eq:#1}}
\newcommand{\refSupp}[1]{Appendix~\ref{supp:#1}}

\newcommand{\mycomment}[1]{}

\definecolor{darkred}{rgb}{0.6,0,0}
\definecolor{darkblue}{rgb}{0.0,0,0.5}
\definecolor{green}{rgb}{0.0,0.5,0}
\definecolor{blue}{rgb}{0,0,0.75}
\definecolor{orange}{rgb}{1,0.6,0.2}
\definecolor{red}{rgb}{1,0,0}
\definecolor{purplish}{rgb}{0.6,0,0.7}

\newcommand{\highlight}[1]{\textcolor{purplish}{#1}}

\renewcommand{\highlight}[1]{#1}

\def\reals{\mathbb{R}}

\def\vdm{f_\theta}
\def\prompt{\mathbf{c}}
\def\encoder{{Enc}}

\def\mesh{\mathcal{M}}
\def\cam{\mathcal{C}}
\def\bg{\mathbf{B}}
\def\tex{\mathcal{T}}
\def\feats{\mathbf{A}}
\def\featsc{\hat{\feats}}
\def\stylized{\mathbf{S}}

\def\mask{\psi}
\def\deff{\tau}
\def\defp{\mathbf{p}}
\def\defpinit{\defp_{\textrm{init}}}
\def\im{\mathbf{I}}
\def\imrgb{\im{}_{rgb}}

\title{MotionDreamer: Exploring Semantic Video Diffusion features for Zero-Shot 3D Mesh Animation}

\author{%
Lukas Uzolas \quad Elmar Eisemann \quad Petr Kellnhofer\\
Delft University of Technology\\
The Netherlands\\
\texttt{\{l.uzolas, e.eisemann, p.kellnhofer\}@tudelft.nl}}

\begin{document}

\maketitle

\begin{abstract}
  Animation techniques bring digital 3D worlds and characters to life.
    However, manual animation is tedious and automated techniques are often specialized to narrow shape classes.
    In our work, we propose a technique for automatic re-animation of various 3D shapes based on a motion prior extracted from a video diffusion model.
    Unlike existing 4D generation methods, we focus solely on the motion, and we leverage an explicit mesh-based representation compatible with existing computer-graphics pipelines.
    Furthermore, our utilization of diffusion features enhances accuracy of our motion fitting.
    We analyze efficacy of these features for animation fitting and we experimentally validate our approach for two different diffusion models and four animation models.
    Finally, we demonstrate that our time-efficient zero-shot method achieves a superior performance re-animating a diverse set of 3D shapes when compared to existing techniques in a user study.
\end{abstract}

\begin{figure*}
    \centering
    
    \includegraphics[width=0.9\textwidth]{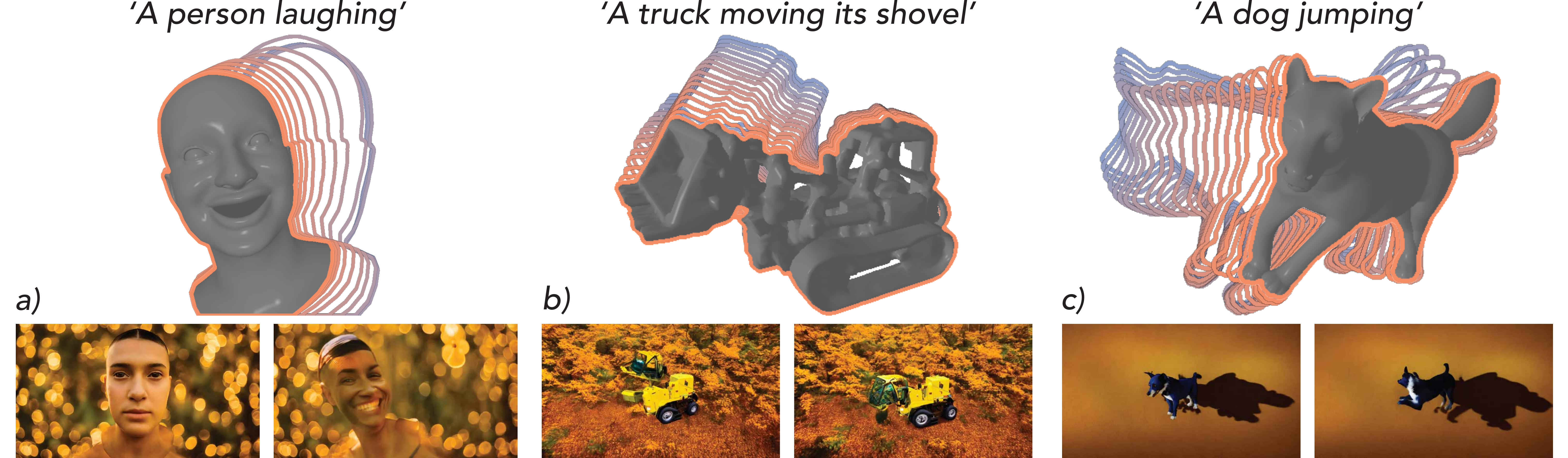}
    \caption{Our Zero-shot 3D mesh animations. 
  From top to bottom: The desired motion description, the resulting animated mesh with motion contours, the driving video from a pre-trained video diffusion model. 
  Notice robustness of our method to the temporal identity shift (a) and to the geometric distortions (b). 
  Diverse shapes are supported through a range of animation models including a) FLAME~\cite{FLAME:SiggraphAsia2017}, b) Neural Jacobian Fields~\cite{aigerman2022njf} and c) SMAL~\cite{zuffi20173d}.}
\label{fig:teaser}
\end{figure*}

%%%%%%%%% BODY TEXT
\section{Introduction}
\label{sec:introduction}
Animation is an important component of video games, simulators, and movies. It makes otherwise rigid environments come to life and is often a result of a tedious motion-data capture coupled to skilled manual editing~\cite{gleicher2000motion}.
However, this does not scale well for applications involving large virtual worlds with thousands of individual entities or for individual objects that are difficult to motion capture due to their physical size or real-world inaccessibility.
For this reason, we propose an end-to-end generative method that re-animates static 3D objects using a pre-trained Video Diffusion Model~\cite{blattmann2023stable,wang2024videocomposer,xing2023dynamicrafter,guo2023animatediff,ho2022imagen} (VDM) without any additional training (\refFig{teaser}).
%However, classical animation either relies on skilled artists~\cite{gleicher2000motion} or used in video games, movies, etc. but animation is hard: It needs expertise~\cite{gleicher2000motion} or deep-learning based methods need big data sets which are either hard to obtain or impossible such as motion data of flies.

We build on the remarkable success of Diffusion models~\cite{ho2020denoising,song2020score}.
Beyond producing nearly photo-realistic 2D images~\cite{nichol2022glide,ramesh2022hierarchical,saharia2022photorealistic,rombach2022high}, diffusion was also adapted for 3D~\cite{poole2022dreamfusion,lee2024text} and 4D shape synthesis~\cite{singer2023text,ling2023align,ren2023dreamgaussian4d,yin20234dgen,wang2023animatabledreamer,bahmani20234d,zheng2023unified,jiang2023consistent4d,zhao2023animate124}.
However, the associated methods suffer from either a high optimization cost and low diversity~\cite{liang2023luciddreamer} of the mode-seeking Stochastic Distillation Sampling~\cite{poole2022dreamfusion} (SDS),
or, as we show in our paper, they are susceptible to the visual artifacts in RGB outputs of existing VDMs.
Furthermore, our method generates a unique animation as a sequence of object poses in a matter of minutes rather than hours common for end-to-end 4D generative methods.
This is a feature crucial for processing of larger sample sets with subsequent filtering based on subjective preferences.
Therefore, we position our approach into a category distinct from end-to-end 4D generation.

%On top of producing nearly photo-realistic images, these models also showcase a surprising versatility in down-stream applications including matching semantic correspondences~\cite{}, predicting monocular depth estimation~\cite{}, or uplifting of 2D images to 3D~\cite{zero123}.
%Diffusion models have been the golden standard for generative tasks in recent years. More interestingly, pre-trained diffusion models can be exploited for a multitude of down-stream task, e.g. semantic correspondence, monocular depth estimation, or 3D-uplifting of 2D images. 

%Similar to 3D uplifting, video diffusion models (VDM) have been used for uplifting static 3D representations to animated 4D sequences. However, previous methods they a) either rely on SDS which is slow and has limited animation diversity due to its mode-seeking nature, or b) are fast but rely to strongly on the pixel output of the video diffusion model which often is broken. Furthermore, none of the previous methods focuses on pre-defined geometry in form of 3D Meshes such that the animations could be used as-is in video game and animation pipelines.

Instead of iterative SDS, we leverage the surprising versatility of semantic features extracted from diffusion models for down-stream tasks such as one-shot segmentation~\cite{khani2023slime} or semantic feature matching~\cite{tang2023emergent, dutt2024diffusion}, which we adapt for motion fitting.
We rely on a classical surface mesh representation in combination with diverse animation models~\cite{aigerman2022njf,SMPL:2015,zuffi20173d,FLAME:SiggraphAsia2017} to obtain animated 3D shapes that are fast to render, compatible with existing rendering frameworks and versatile across object classes.

In summary, we present the following contributions:
1. We introduce a novel zero-shot generative method for 3D mesh animation based on rendering in the semantic feature space of pre-trained VDMs.
% matching \lukas{/rendering?} semantic features of pre-trained VDMs.
2. We analyze effectiveness of VDM features for pose estimation to validate our method and design choices.
3. We evaluate two different VDMs and four animation models and demonstrate a preference of our 3D animations to existing generative approaches in a user study.
%We demonstrate comparable results to state-of-the-art methods in the text-to-4D setting while being significantly faster and more robust (?). 
%3. We show that our approach is robust w.r.t. type of meshes, deformation models, and choice of video diffusion model. We further point out that current state-of-the-art VDMs are still lacking in terms of textural control over the motion.

\section{Related Work}

\label{sec:related}

Our method exploits VDMs to create novel animations of 3D objects.
Here we discuss relevant work on video generation and existing approaches for 3D shape representation and animation.

\subsection{Video generation}

Generative visual models have advanced rapidly from Variational Auto-Encoders~\cite{kingma2013auto}, Normalizing Flows~\cite{dinh2014nice,rezende2015variational} and Generative Adversarial Models~\cite{goodfellow2014generative} to Diffusion Models~\cite{ho2020denoising,song2020score} and Continuous Normalizing Flows~\cite{lipman2022flow} achieving a nearly photorealistic image synthesis~\cite{nichol2022glide,ramesh2022hierarchical,saharia2022photorealistic,rombach2022high} as well as state-of-the-art video synthesis~\cite{blattmann2023stable,wang2024videocomposer,xing2023dynamicrafter,guo2023animatediff,ho2022imagen}.
Surprisingly, the features learned by the U-Net~\cite{ronneberger2015u} of many diffusion models exhibit semantic properties useful for down-stream tasks such as segmentation~\cite{khani2023slime} and feature matching~\cite{zhang2024tale, tang2023emergent, luo2024diffusion}.
Consequently, we analyze utility of two such models~\protect\cite{wang2024videocomposer,xing2023dynamicrafter} for our motion fitting, while we leave  opportunities presented by recent large VDMs~\protect\cite{videoworldsimulators2024,googledeepmind2024veo} utilizing Visual Transformers~\protect\cite{dosovitskiy2020image} as an avenue for future research.

\subsection{Shape and pose representations}
There exist many ways for representing 3D shapes from classical explicit representations including point-clouds, voxels or surface meshes favored in real-time applications, to implicit neural shape representations~\cite{mildenhall2020nerf,park2019deepsdf,tewari2022advances} enabling photorealistic 3D scene reconstruction. 
In the middle, 3D Gaussians~\cite{kerbl20233d} have been shown to combine advantages of both at an increased storage cost.
In this paper we focus on surface meshes for their fast rendering, efficient storage and wide application support. 

While animation of object poses
can be encoded as a sequence of static representations~\cite{maglo20153d}, specialized representations ease editing for both arbitrary and class-specific shapes. In the first category, deformation fields offer maximal flexibility for dense volumetric optimization~\cite{pumarola2020d}, Neural Jacobian Fields (NJFs)~\cite{aigerman2022njf} offer space-time continuity and smoothness for surface optimization and external cages reduce the control space for easier editing~\cite{xu2022deforming}.
In the second category, low-dimensional templates support manual animation and motion capture
by combining Linear Blend Skinning~\cite{lewis2000pose} and Blend Shapes~\cite{parke1972computer,parke1974parametric}
for specific classes of shapes such as
faces~\cite{Blanz:99,FLAME:SiggraphAsia2017}, bodies~\cite{SMPL:2015,SMPL-X:2019}, hands~\cite{MANO:SIGGRAPHASIA:2017}, or even animals~\cite{zuffi20173d}.
Our method is agnostic to the choice of an animation model, which we test on high-dimensional NJFs~\cite{aigerman2022njf} and on low-dimensional templates~\cite{SMPL:2015,zuffi20173d,FLAME:SiggraphAsia2017}.

\subsection{3D motion and animation}

\paragraph{Capture}
Motion, most often for humans, can be directly captured~\cite{o1980model} using sparse inertial sensors~\cite{tautges2011motion} or dense visual observations~\cite{hogg1983model} either with tracking markers~\cite{sigal2010humaneva} or without them~\cite{sigal2004tracking}.
For a monocular video, we can estimate 2D poses~\cite{pishchulin2012articulated,toshev2014deeppose,cao2017realtime} and uplift them to 3D~\cite{Bogo:ECCV:2016,martinez2017simple,lutz2022jointformer,zhang2024rohm,shin2023wham} thanks to data priors~\cite{ponton2023sparseposer,SMPL-X:2019} based on large motion datasets~\cite{ionescu2013human3,joo2015panoptic,mahmood2019amass}.
%While similar models exists for faces~\cite{DECA:Siggraph2021} or hands~\cite{}, the class-specific data requirement hampers wider generalization.
However, the specific training for each class limits generalization.
In contrast, recent advances in neural rendering~\cite{mildenhall2020nerf,kerbl20233d,tewari2022advances} enabled class-agnostic 4D reposable reconstructions~\cite{noguchi2022watch,yao2022hi,uzolas2024template}.
Our method is similarly based on class-agnostic differentiable pose optimization but differently from a direct image supervision, we exploit diffusion features of a monocular video rather than multi-view observations.
    
\paragraph{Generation}
Learned priors can also be used for text-conditioned motion synthesis~\cite{zhu2023human}.
However, this is in practice limited to human domain~\cite{guo2022generating,tevet2022motionclip,jiang2024motiongpt,tevet2023human} where annotated 3D motion datasets exist~\cite{Plappert2016,Guo_2022_CVPR} or to other skeletal shapes~\cite{kapon2023mas} if at least 2D annotations are available.
Alternatively, image and video generative models enabled class-agnostic joint shape and motion 4D generation~\cite{singer2023text,ling2023align,ren2023dreamgaussian4d,yin20234dgen,wang2023animatabledreamer,bahmani20234d,zheng2023unified,jiang2023consistent4d,zhao2023animate124} is usually based on Stochastic Distillation Sampling (SDS)~\cite{poole2022dreamfusion} which, however, narrows the sampled distribution~\cite{liang2023luciddreamer} due it its mode-seeking behavior.
Closest to us, \citet{ren2023dreamgaussian4d} extract motion from a full video input.
Our method shares the idea of extracting motion from a video model but thanks to utilizing the feature space it produces more natural motion with fewer visual artifacts.
Furthermore, we do not use 3D uplifting methods requiring background masks such as Zero-1-to-3 \cite{liu2023zero}.
Additionally, by focusing on motion alone we achieve faster sampling.
Finally, both captured or generated motion can be transferred from one shape to another~\cite{gleicher1998retargetting}, either based on morphological similarity~\cite{liao2022skeleton,wang2023zero} or data-driven domain matching~\cite{rhodin2014interactive,li2023ace}. 
We experimentally show that our method is preferable when neither of the two conditions can be satisfied.

\section{Preliminaries}
Our method exploits internal representation of VDMs. 
Here, we provide a brief summary of these models and semantic information encoded in their internal features.

\label{sec:preliminaries}

\subsection{Video Diffusion Models}
VDMs are a type of a generative model producing video sequences by gradual denoising~\cite{song2020score,ho2020denoising} of a Gaussian-noise image sequence $\mathbf{z} \in \mathbb{R}^{L\times H \times W \times D_\textrm{lat}}$, where $L$ is the frame count, $H, W$ the spatial dimensions, and $D_\textrm{lat}$ is 3 for RGB models or the latent feature dimension for Latent Diffusion~\cite{rombach2022high}.
The forward diffusion process $q(\mathbf{z}_t|\mathbf{z}_0,t)$ gradually transports $\mathbf{z_0} \equiv \mathbf{z}$ to the Gaussian-noise prior over $T$ steps such that $\mathbf{z}_T \sim \mathcal{N}(\mathbf{0}, \mathbf{I})$.
This is used to learn a denoising function $\vdm(\mathbf{z}_t,t,\prompt)$ as a $\theta$-parameterized network approximating the reverse process $p_\theta(\mathbf{z}_{t-1}|\mathbf{z}_t, t, \prompt)$.
A commonly used $\epsilon$-prediction training procedure minimizes an objective
$\sum_{t,\prompt, \mathbf{z}, \epsilon} \left\| \epsilon - \vdm(\mathbf{z}_{t}, t, \prompt) \right\|^{2}_{2}$
across data and noise samples $\mathbf{z} \sim p_{\text{data}}$ and $\epsilon \sim \mathcal{N}(\mathbf{0},\mathbf{I})$.
Finally, sampling the noise prior $\mathcal{N}(\mathbf{0}, \mathbf{I})$ and reversing the diffusion yields video generation.
The conditioning vector $\prompt \in \mathbb{R}^N$ is often a text embedding, image embedding or both, and it steers the process, often with a classifier-free guidance~\cite{ho2021classifier}.

\subsection{Semantic Diffusion Features}
\label{sec:feat_extraction}
Intermediate activations of image diffusion networks have been shown to encode semantic features and provide robust correspondences across image samples~\cite{zhang2024tale, tang2023emergent, luo2024diffusion}. 
We adopt the methodology of~\citet{tang2023emergent}, where $\vdm$ is parameterized by a U-Net. 
The semantic feature maps $\feats_u \in \mathbb{R}^{H_u \times W_u \times A_u}$ are extracted from the intermediate activations of a U-Net layer $u$ with height, width and feature size $H_u$, $W_u$, and $A_u$.

Given a pair of images with feature maps $\feats_u, \mathbf{B}_u$ and a chosen spatial location $\mathbf{\phi}^A \in \reals^2$ in the first image,
we  find a semantically corresponding spatial location $\mathbf{\phi}^B \in \reals^2$ in the other image as
$
    \mathbf{\phi}^B = 
    \mathop{\mathrm{arg\,max}}_{\mathbf{\phi}^B}     
    \kappa(\feats_u[\phi^A], \mathbf{B}_u[\phi^B])$,    
%\end{equation}
where
\begin{equation}
\label{eq:finding_correspondence}
    \kappa(\mathbf{a}, \mathbf{b}) =  
    \frac{
        \mathbf{a}^\textrm{T}
        \mathbf{b}
    }{
        ||\mathbf{a}||_2
        ||\mathbf{b}||_2
    }
\end{equation}
is a cosine similarity $\kappa: \reals^{A_u}\times\reals^{A_u}\to\reals$ and
$\mathbf{x}[\phi]$ denotes spatial sampling of a map $\mathbf{x}$ at location $\phi$, which we implement as a bilinear interpolation.
For video, we treat each frame as an image with its own feature map, and we
optimize semantic correspondences of reposed meshes using \refEq{finding_correspondence}.

\section{Method}
\label{sec:method}
\begin{figure*}[]
\centering
  \includegraphics[width=0.9\textwidth]{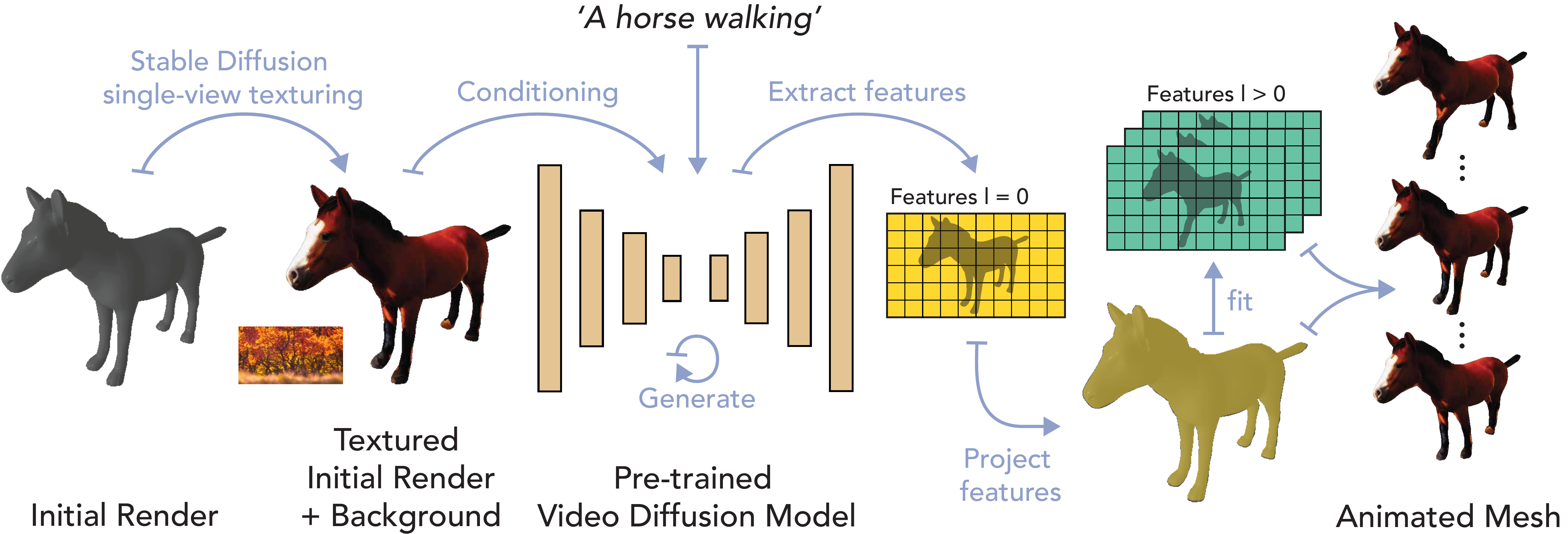}
  \caption{
  A diagram of our method. 
  First, we automatically texture the input mesh $\mesh$ to reduce the domain gap to the VDM prior (\refSec{scene_init}).
  Second, we condition the VDM by a rendered image $\imrgb$ to produce a video with motion and to extract features $\featsc$ for all $L$ frames from its internal U-Net (\refSec{motion_generation}).
  Finally, we reproject the input frame features $\featsc^0$ on the mesh surface and we optimize mesh animation parameters $\defp$ to match the reposed mesh features to the video (\refSec{optimization}).
  }
  \label{fig:method}
\end{figure*}

Our methods accepts an unseen 2-manifold 3D mesh in an arbitrary pose and uses a pre-trained VDM to generate a temporal sequence of animation parameters (see~\refFig{method} for an overview).
We first describe our method for a general VDM and animation model before discussing specific realizations in \refSec{impl}.

\paragraph{Definitions}
We define the input mesh $\mesh$ as a tuple of $N$ vertices and $M$ triangular faces  $\mesh{} := (\{\mathbf{u}_n \in \reals^3 | n = 0,...,N-1\}, \{\mathbf{f}_m \in \mathbb{N}^3| m = 0,...,M-1\})$. 
Next, we define $\deff: (\mesh{}, \defp{}) \rightarrow \mesh^\prime$ as a function transforming vertices to produce a mesh $\mesh^\prime := (\{\mathbf{u}_n^\prime\}, \{\mathbf{f}_m\})$ with a novel pose described by animation parameters $\defp{} \in \reals^P$.
We refer to $\defpinit$ as the input pose where $\deff(\mesh, \defpinit) \equiv \mesh$ and, without a loss of generality, we assume it matches the first frame.
Finally, $r_{rgb}: (\mesh{}, \cam{}, \tex, \bg) \rightarrow \imrgb{}$ is a rendering function producing an RGB image $\imrgb{} \in \reals^{H\times W \times 3}$ of the mesh $\mesh{}$ for a \highlight{manually defined canonical} camera $\cam{}$, surface texture $\tex$, and a background image $\bg \in \reals^{H\times W \times 3}$.

\subsection{Single-View Texturing}
\label{sec:scene_init}
\label{sec:single_view_texturing}
While the visual datasets used to train existing VDMs are very large, they favor natural looking textured images with backgrounds (see \refSupp{texturing_effect} for examples).
We reduce the domain gap for our rendered image by automatically generating an RGB texture $\tex$ and a semantically fitting background image $\bg{}$.
First, we render a depth map and a foreground mask $\mask$ for a single fixed viewpoint of $\mesh$.
Next, we style-transfer the depth map using a pre-trained ControlNet diffusion model~\cite{zhang2023adding} conditioned by a user-provided textual description to obtain a textured RGB image $\stylized$.
Then, we crop the foreground texture $\tex = \textit{unproject}(\stylized \odot \mask)$ and apply it to the mesh $\mesh_0$ using projective texturing~\cite{williams1978casting}.
Importantly, we do not strive for a complete texture of the entire mesh, but merely for a stylization of the single-view VDM input image.
Finally, we obtain the background image $\bg{}$ by inpainting 
the remainder of $\stylized$ outside of the foreground bounding box
using Stable Diffusion XL~\cite{podell2023sdxl}. See \refSupp{stimuli} for prompt details.

\subsection{Motion Generation}
\label{sec:motion_generation}
The motion produced by our method originates from a VDM conditioned 
by our rendered mesh image $\imrgb^0 = r_{rgb}(\mesh{}^0, \cam{}, \tex, \bg)$ and an embedding of the intended motion text description.
We sample the generator in a multi-step diffusion process over $T$ steps denoted as $t \in [0,...,T-1]$ with scheduling details specific to each VDM. 
Because the temporally incoherent visual artifacts in RGB video outputs make motion tracking difficult (see \refFig{teaser}),
we extract semantically meaningful U-Net features $\feats^t_u$ at time step $t = \hat{t}$ and U-Net layer $u = \hat{u}$ as explained in \refSec{feat_extraction},
and we show that this improves the fitting accuracy.
We motivate our choice of $\hat{t}$ and $\hat{u}$ in \refSec{ablation}, and will omit the suffixes from now on for brevity, such that $\featsc \in \reals^{L \times \hat{H} \times \hat{W} \times \hat{A}} \equiv \feats^{\hat{t}}_{\hat{u}}$ and $\featsc^l$ selects the video frame $l$ of $L$.
We further assume $\featsc^0$ corresponds to the input image $\imrgb^0$ (see \refSupp{vdms} for a discussion).

\subsection{Motion fitting}
Given the known correspondence of the mesh $\mesh$, initial pose $\defpinit$, image $\imrgb^0$ and features $\featsc^0$ for the input frame $l=0$, we aim to recover all animation parameters $\defp^l$ for $l \in [0,...,L-1]$.
We achieve this by optimizing $\defp$ to match reprojections of the input $\featsc^0$ to $\featsc^l$ extracted from the video.
To this goal, we first reproject $\featsc^0$ to new poses $\defp^l$ and optimize these poses using a gradient descent.

\paragraph{Feature Reprojection}
\label{sec:feature_mesh}
Our mesh pose fitting is based on reprojection of $\featsc^0$ to any new pose $\defp^l$. 
First, we use projective texturing to map $\featsc^0$ to $\mesh$.
We obtain per-vertex features $\{\mathbf{a}_n\}$ by mapping each mesh vertex $\mathbf{u}_n$ to the image plane of the camera $\cam$ and sampling $\featsc^0$ as $
%\begin{equation}
    \mathbf{a}_n = \featsc^0[P(\mathbf{u}_n,\cam)],
%\end{equation}
$ where $P(.)$ is a world space to image plane projection function and $[.]$ is a bilinear sampler. 
Finally, we transform $\mesh$ to $\mesh^l = \deff(\mesh, \defp^l)$ for a given novel pose $\defp^l$ and we render a feature image 
\begin{equation}
\label{eq:imrf}
\im{}_{\feats}^l = r_{\feats}(\mesh{}^l, \cam{}, \{\mathbf{a}_n\}, \bg_\feats)
\end{equation}
where $r_{\feats}$ is a rasterization function interpolating the vertex attributes $\{\mathbf{a}_n\}$ and $\bg_\feats$ is a background feature map produced by inpainting the background $\featsc^0 \odot (1-\mask)$ with a mean of valid features.
Notice that \refEq{imrf} implies an approximate identity $\im{}_{\feats}^0 \approx \featsc^0$, and we optimize $\defp$ to improve this match for the full animation.

\paragraph{Mesh Pose Optimization}
\label{sec:optimization}
We observe that direct optimization of each $\defp{}^l$ independently is prone to local minima.
Instead, we exploit the implicit bias of Multi-Layer-Perceptrons (MLPs) towards smooth functions, and regress $\defp{}^l$ as a frame-dependent offset from an initial pose $\defpinit$ such that $\defp{}^l = \alpha m_{\omega}(\gamma(l)) + \defpinit$, where $\alpha=0.01$ is a scaling constant, $\gamma$ is a frequency encoding \cite{mildenhall2020nerf}, and $m(.)$ is an MLP with learnable parameters $\omega$.
We optimize $\omega$ by gradient descent to enforce semantic correspondences between the animated mesh and the video, i.e. $\im{}_{feat}^l \approx \featsc^l$, using the rendering loss: 
\begin{equation}
    \mathcal{L}_{r} =
    1 -
    \frac{1}{L \hat{H} \hat{W}} 
    \sum_{l=0}^{L-1} 
    \sum_{i \in \Omega_\feats} 
        \kappa(\im_{feat}^l[i], \featsc^l[i]),
\end{equation}
where $\kappa()$ is the cosine similarity (\refEq{finding_correspondence}),
$\Omega_\feats$ is the spatial domain of $\featsc$ and $[i]$ a spatial sampler.

\paragraph{Regularization losses} 

First, our monocular video provides only a limited supervision for motion-in-depth.
We discourage the optimization from explaining spatial deformation artifacts in the input video via motion-in-depth by per-vertex regularization loss

\begin{equation}
\mathcal{L}_{d} = \frac{1}{L N} \sum_{l=0}^{L-1} \sum_{n=0}^{N-1} || (\bar{d}^0 - d^0_n) - (\bar{d}^l - d^l_n) ||_1,
\end{equation}
where $d^l_n$ is the projected depth of vertex $u_n$ in frame $l$, and $\bar{d}^l = 1/N \sum_{n=0}^{N-1} d^l_n$. 
Second, we enforce temporal smoothness beyond the MLP's implicit bias to further reduce jitter using the smoothness loss $\mathcal{L}_{s} = 1/((L-1)N) \sum_{l=0}^{L-2} ||\defp{}^l - \defp{}^{l+1}||_1$. 
Lastly, we penalize propagation of local spatial distortions from video by suppressing large deformations using the fidelity loss $\mathcal{L}_{f} =  1/(LN) \sum_{l=0}^{L-1} ||\defp{}^l||_1$. 
Consequently, our complete optimization objective is $\mathcal{L} = w_r\mathcal{L}_{r} + w_d\mathcal{L}_{d}  + w_s\mathcal{L}_{s} + w_f\mathcal{L}_{f}$ with $w_r = 5$, $w_d = 0.01$, $w_s = 0.1$, $w_f = 0.01$.

\subsection{Implementations Details}
\label{sec:impl}
We implement our method in PyTorch~\cite{paszke2017automatic} with PyTorch3D~\cite{ravi2020pytorch3d} mesh rasterizer, and we optimize the poses with the Adam optimizer~\cite{kingma2014adam} for $1\,000$\,steps.
We discuss further details in \refSupp{impl_details}.

\paragraph{Animation Models}
We experiment with four animation models for poses $\defp$.
For domain specific shapes, we use low-dimensional articulated models SMPL~\cite{SMPL:2015} (for humans), SMAL~\cite{zuffi20173d} (animals) and FLAME~\cite{FLAME:SiggraphAsia2017} (faces), where $\defp^l$ are the joint angles and the other shape parameters are fixed.
For other meshes, we use Neural Jacobian Fields (NJF)~\cite{aigerman2022njf}, which encodes the pose $\defp^l$ by surface Jacobians, in combination with a single global translation and rotation -
see \refSupp{anim_models} for details and for an additional rigidity regularizer $\mathcal{L}_{j}$ applied for NJF.

\paragraph{VDMs}
We evaluate 2 VDMs: VideoComposer~\cite{wang2024videocomposer} (VC) and DynamiCrafter~\cite{xing2023dynamicrafter} (DC) with $\featsc$ resolution of $(160, 88)$ and $(128, 72)$ respectively ($1/8$ of their outputs). We use their recommended inference schedulers with $T=50$\,steps.
Our assumption of $\featsc^0 \sim \imrgb^0$ is satisfied by design for VC, and we present a solution for DC in \refSupp{vdms}.
We empirically find VC performs better for images with the background $\bg$, while DC performs well even with a uniform white background. 
\highlight{Additionally, we assessed another VDM, Stable Video Diffusion~\cite{blattmann2023stable}, but we discarded due to its low motion quality (see Appendix \ref{supp:texturing_effect}, \ref{supp:svd_results})}.

\section{Experiments}
\label{sec:validation}
We compare our zero-shot motion generation to other methods in a user study. 
Further, we quantitatively evaluate our pose fitting algorithm on a synthetic human motion dataset and measure the contribution of the individual components in an ablation study.

\begin{figure*}
    \centering
    \includegraphics[width=0.81\linewidth]{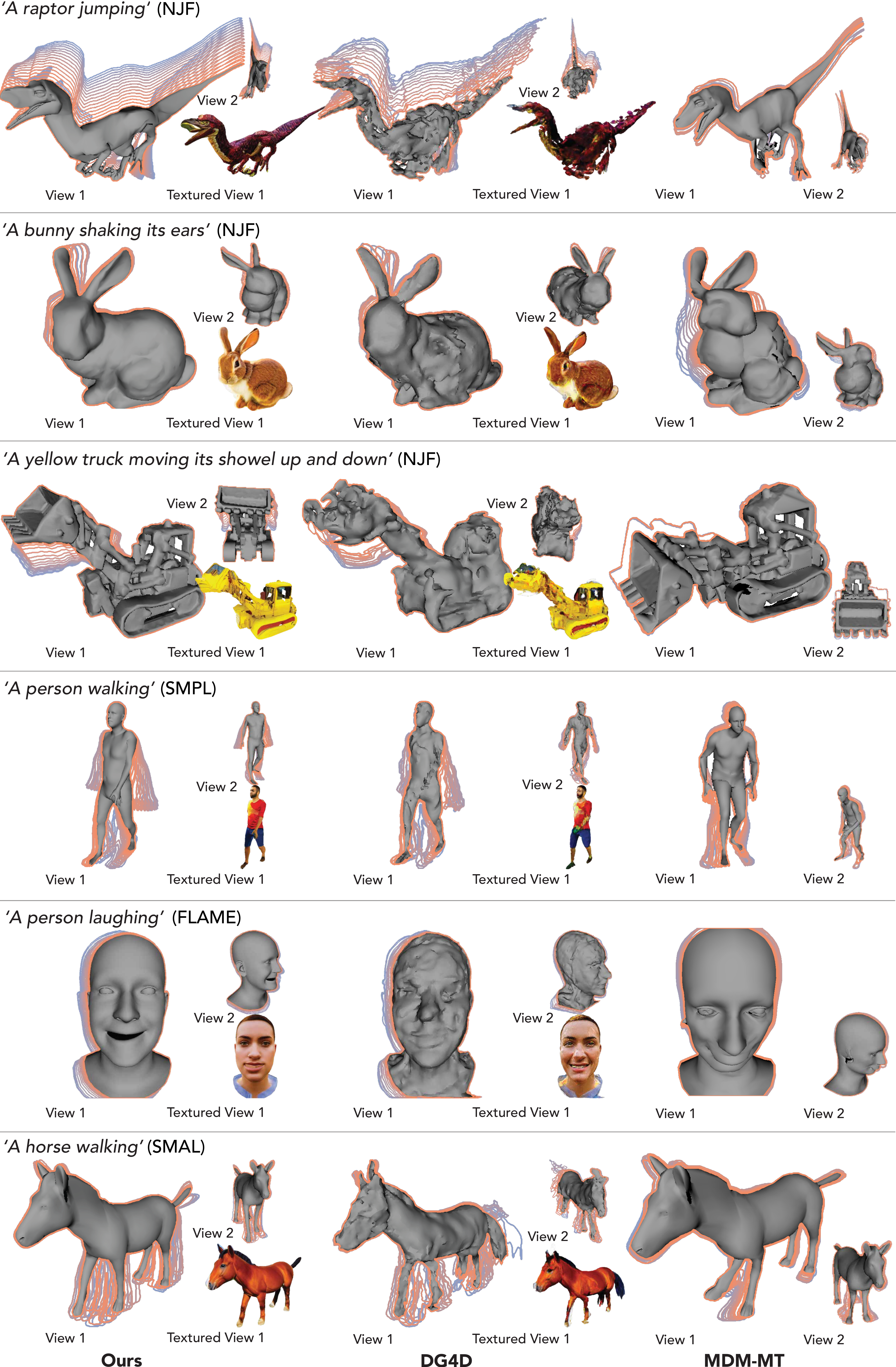}
    \caption{
    A qualitative comparison of our method to DG4D and MDM-MT for the prompts and the shapes used in our study. 
    We display 2 untextured views of the last frame with one one additional textured image for reference.
    The contours convey the motion trajectory.}
    \label{fig:qualitative_results}
\end{figure*}

\subsection{User Study}
\label{sec:user_study}
We compare our method to two other approaches for zero-shot 3D motion synthesis. First, we compare to DG4D~\cite{ren2023dreamgaussian4d} as an end-to-end shape-and-motion generative method based on image and video diffusion.
Second, in absence of a class-agnotic method, we compare to a human motion diffusion model (MDM)~\cite{tevet2023human} combined with motion retargeting (MT)~\cite{liao2022skeleton}.
\highlight{We provide an additional qualitative comparison to a contemporary end-to-end generative method Consistent4D~\cite{jiang2023consistent4d} in \refSupp{consistent4d}.}
We run our method with both VC and DC backbones and use the same generated videos as inputs for DG4D (see \refSupp{baselines} for details).
We use 9 meshes and a total of 12 prompts combined to obtain 2 human stimuli (using the SMPL mesh), 2 horses (SMAL), 2 faces (FLAME) and 4 other stimuli each with a unique mesh (NJF).
See~\refFig{teaser}, \refFig{qualitative_results}, and the supplemental video for visual examples, \refSupp{stimuli} for a complete list.

12 participants aged 24--41, na\"ive to the purpose of the experiment and with a normal or corrected-to-normal vision,  participated in a $\sim20$\,min low-risk IRB approved study, after signing an informed consent without any compensation.
16 frame (1 second) long video pairs from different methods were presented side-by-side in random order.
Each displayed the same untextured animated shape from two viewpoints to clearly display the motion.
Videos were looped until a binary answer was entered using a keyboard.
The same stimuli were used for three different questions in three blocks.
See \refSupp{study_instructions} for details.

\begin{figure*}
    \centering
    \includegraphics[width=\textwidth]{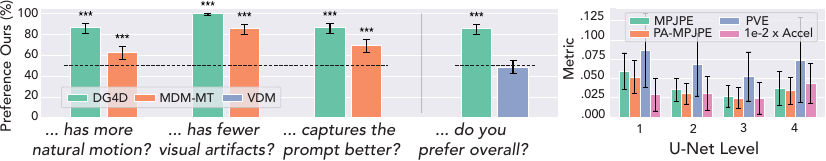}
    \caption{
    Left: Results of the user study, asking the question: \textit{"Which video... ?"} 
    For the first three questions we compare our method against untextured renders of DG4D and MDM-MT. 
    For the last question we compare against the full RGB outputs of DG4D and the VDM output. 
    *** denotes significance at $p < 0.001$ (bars show 95\% confidence intervals). 
    Right: Pose fitting errors for $\feats^{\hat{t}}_u$ extracted across U-Net layers $u$ with bars showing standard deviations.
    }
    \label{fig:user_study}
\end{figure*}

In \refFig{user_study} (Left), we observe a statistically significant preference for our method compared to both DG4D and MDM-MT in terms of having ``more natural motion'', ``fewer visual artifacts'' and ``capturing the prompt better'' ($p < 0.001$, binomial test).
We provide a break-down for individual shapes in \refSupp{study_results}.
As expected, the human-specific MDM-MT approach excels for human stimuli but fails for morphologically distinct shapes, where correspondences are difficult to establish, which results in semantically incorrect and visually distracting motion (see \refFig{qualitative_results} ``Bunny'').
In contrast, the other class-agnostic model, DG4D, struggles to accurately represent the video motion sequences leading to noisy reconstructions (see \refFig{qualitative_results} ``Raptor''). 
Moreover, the motion optimization in DG4D (Stage 2) takes $233 \pm 5$ seconds on an NVIDIA RTX 3090, while our method leverages fast rasterization and performs pose optimization in only $148 \pm 39$ seconds.
\refFig{teaser}, \refFig{qualitative_results}, \refSupp{study_results}, and our video show more examples.

\paragraph{Input Accuracy}
After the main study, we asked each participant to additionally compare our textured output to the full DG4D color rendering and to the unprocessed VDM videos for overall preference (see the last bars \refFig{user_study} Left).
The participants strongly prefer our method to DG4D, likely due to the more accurate geometry (\refFig{qualitative_results}).
There was no effect when comparing to the VDMs, suggesting that our method closely preserves characters of the generated videos and should, therefore, benefit from future VDMs with a more accurate motion depiction.

\subsection{Pose Optimization}
\label{sec:pose_est}
We observe that the limiting factor of our method is the VDM motion quality.
To remove this influence, we quantitatively evaluate performance of our pose fitting component using a captured human dancing motion dataset AIST++~\cite{li2021ai} with known poses.
First, we randomly select 20 test sequences and re-render the first $20$ frames from each using the available SMPL mesh to simulate a perfect VDM.
Then, we use VC to extract $\featsc$ from the rendered video following \citet{tang2023emergent} and optimize $\defp$ for the SMPL model (\refSec{optimization}) before evaluating the common metrics~\cite{shin2023wham}: the Mean Per Joint Position Error (MPJPE), the Procrustes-aligned MPJPE (PA-MPJPE), the Per Vertex Error (PVE), and finally the Acceleration error (Accel) for smoothness.

\begin{figure}
    \centering
    \includegraphics[width=\linewidth]{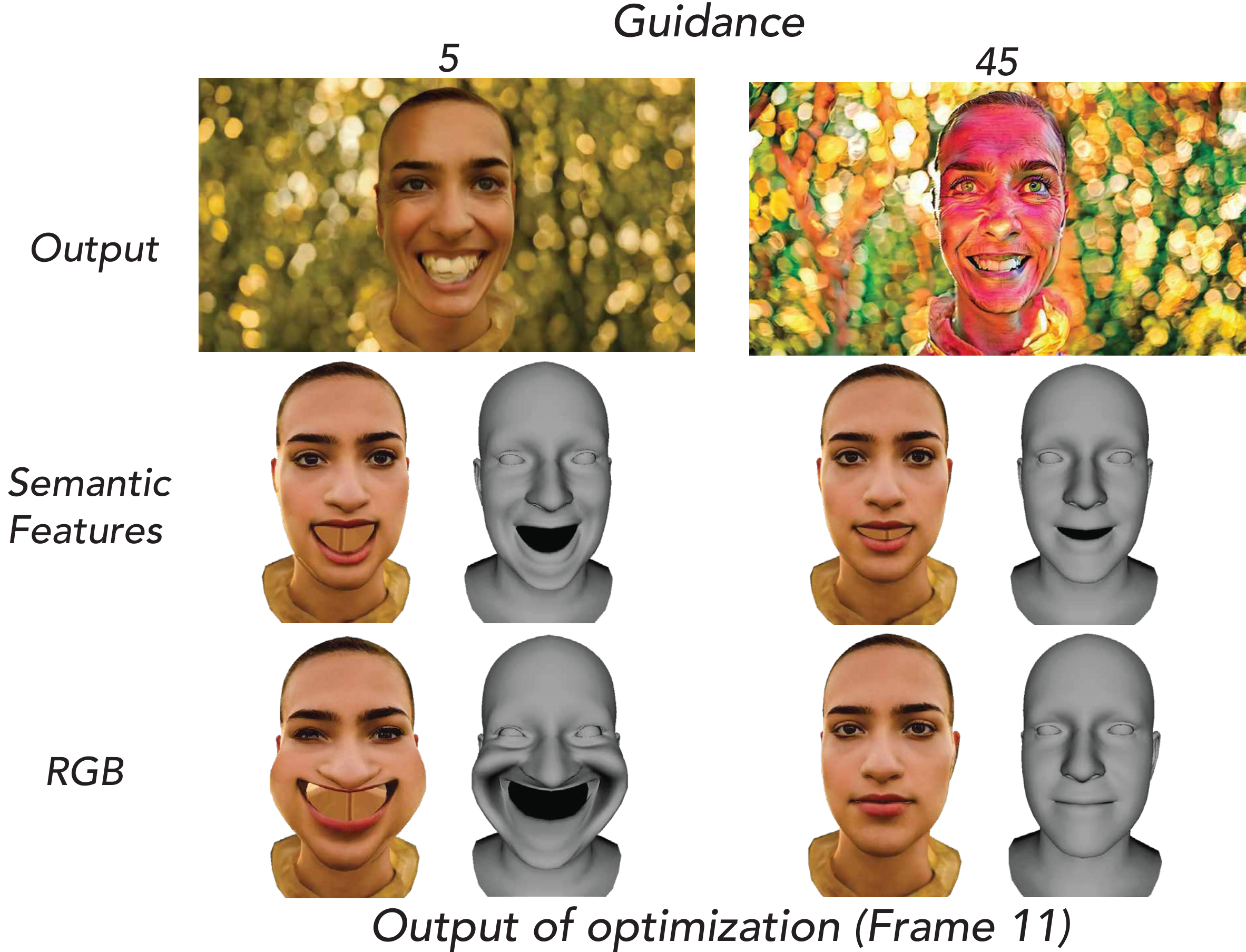}
    \caption{Effect of features on optimization under different guidance scales. Higher guidance scales are usually accompanied by stronger artifacts in RGB space (right column). The top row shows frame 11 of the VDM output, i.e. the target expression. The consecutive rows show semantic features vs RGB used for optimization.}
    \label{fig:ablation_features}

\end{figure}

We conduct three comparisons. 
First, we compare our single-view texturing (\textit{textured}, \refSec{single_view_texturing}) to a uniform gray shading (\textit{untextured}).
Second, we compare our semantic features $\featsc$ (\textit{Ours}) to RGB features (\textit{RGB}) extracted directly from the input videos. 
Finally, we additionally test a state-of-the-art human pose estimation method WHAM~\cite{shin2023wham} as a domain-specific reference.
Since our method always starts with known $\defpinit$, we emulate the same for WHAM by measuring its 
first-frame per-joint error and transform all predictions accordingly.
This empirically improves WHAM scores relative to the unprocessed outputs.
\refSupp{pose_est_study} provides details and alternative alignment strategies.
\paragraph{Results}
As summarized in \refTbl{pose_fitting}, \textit{Ours} consistently achieves better results with \textit{textured} inputs than with \textit{untextured} inputs, which motivates our Single-View Texturing (\refSec{single_view_texturing}).
Furthermore, \textit{Ours (full)} with semantic features $\featsc$ achieves lower errors than the variant with \textit{RGB} features, which documents the utility of these features for our task.
Moreover, \textit{Ours (full)} compares favorably even to the WHAM pose estimator despite the lack of human-specific training.
This might be explained by the artificial appearance of our input videos which differ from common human pose estimation datasets. We do not claim general supremacy of our method for human pose estimation.
This is showcased in \refFig{failure_cases} (right), where our method struggles to avoid physiologically implausible poses.
\highlight{Finally, in \refFig{ablation_features} we compare both features qualitatively in our full generative method and confirm that our semantic features lead to a better motion fit with fewer artifacts.
See \refSupp{rgb_vs_semantic_features} for more examples.}

\begin{table}[t!]
\footnotesize
\begin{tabular}{lcccc}
\hline
    & \textbf{MPJPE} & \textbf{PA-MPJPE} & \textbf{PVE}   & \textbf{Accel} \\ \hline
\multicolumn{5}{l}{\textbf{\textit{Textured} (default)}}                         \\ \hline 
WHAM  & .059 ± .029 & .042 ± .016   & .075 ± .036 & 7.9 ± 9.0 \\ % copying gt root R and t
RGB    & .044 ± .051 & .044 ± .042   & .077 ± .059 & 7.5 ± 16.7 \\
\textbf{Ours (full)}   & \textbf{.041} ± .036 & \textbf{.039} ± .035    & \textbf{.063} ± .057 & \textbf{5.0} ± 7.2 \\ \hline
\multicolumn{5}{l}{\textit{Untextured}}                         \\ \hline 
WHAM & .057 ± .028 & \textbf{.039} ± .015    & \textbf{.070 }± .035 & 7.4 ± 9.1 \\ % copying gt root R and t, learn body transform
RGB &   .146 ± .056 & .126 ± .043    & .203 ± .074 & \textbf{3.2} ± 3.0 \\
Ours & \textbf{.051} ± .037 & .044 ± .034  & .073 ± .054 & 4.7 ± 5.7
\end{tabular}

\caption{
The pose fitting performance of WHAM~\cite{shin2023wham} and variants of our method for re-rendered AIST++ human body sequences~\cite{li2021ai}. 
Less is better for all metrics (see~\protect\refSec{pose_est}). }

\label{tbl:pose_fitting}
\end{table}

\begin{table}[t!]
\begin{minipage}[t]{0.38\linewidth}

\includegraphics[width=\linewidth]{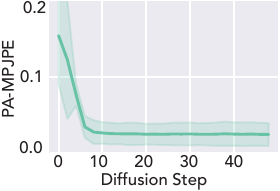}

\end{minipage}
\hfill
\begin{minipage}[t]{0.62\linewidth}
\vspace{-2cm}
\includegraphics[width=\linewidth]{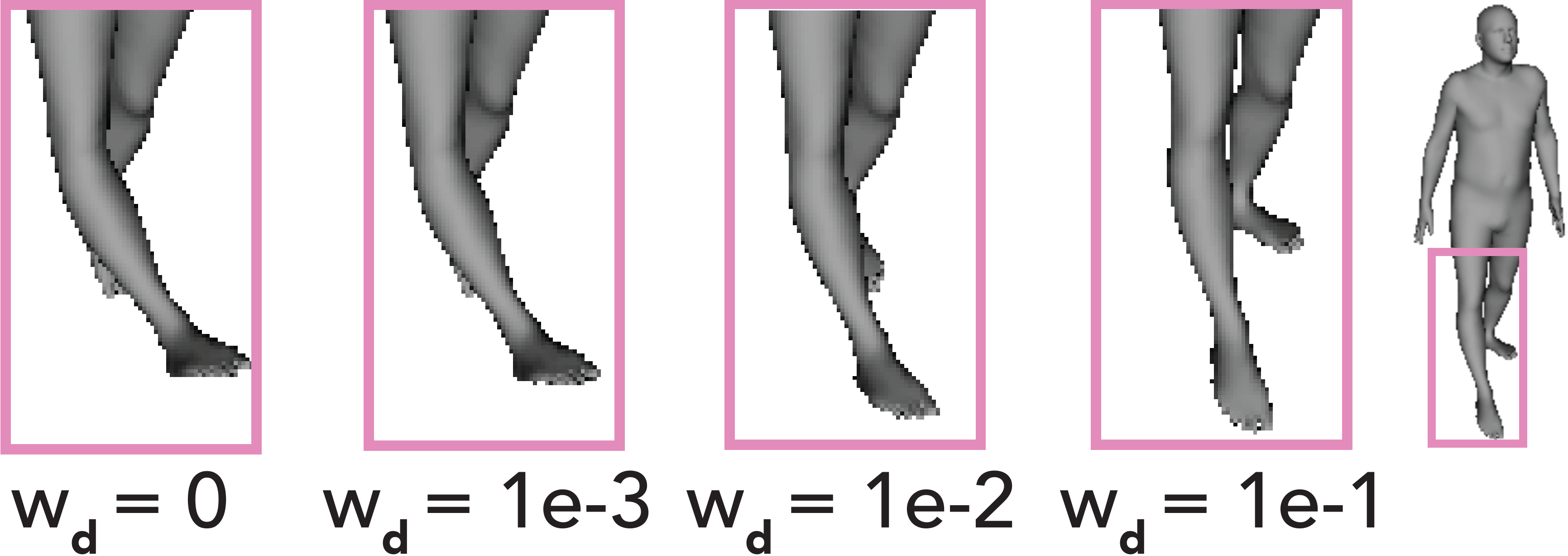}
\end{minipage}
\captionof{figure}{
Left: PA-MPJPE~$\downarrow$ with a standard deviation range for features $\feats^t_{\hat{u}}$ extracted for different diffusion steps $t$. Right: Depth regularization prevents undesirable motion-in-depth explanations.
}
\label{fig:diffusion_timestep}
\label{fig:losses}
\end{table}

\begin{table}[]
\footnotesize
\begin{tabular}{lcccc}
\hline
        & \textbf{MPJPE}        & \textbf{PA-MPJPE}     & \textbf{PVE}          & \textbf{Accel}       \\ \hline
no $\mathcal{L}_s$ & .027 ± .016 & .025 ± .016          & .053 ± .036          & \textcolor{red}{\textbf{2.72 ± 2.19}} \\
no $\mathcal{L}_f$ & .026 ± .017          & .025 ± .018 &  \textcolor{red}{\textbf{.103 ± .044}}  & 2.52 ± 2.28          \\
no $\mathcal{L}_d$ & .025 ± .006          & .024 ± .008          & .046 ± .016          & 2.46 ± 2.04          \\
Full    & .027 ± .016 & .025 ± .016          & .053 ± .036          & 2.50 ± 2.36         
\end{tabular}
\caption{
Performance of our ablated method variants in pose fitting. Notable performance impacts highlighted in red.
}
\label{tbl:losses}
\end{table}

\subsection{Ablations}
\label{sec:ablation}

We reuse the pose optimization experiment to validate our design choices. 
To this end, we follow the same procedure for 6 of the same AIST++ sequences~\cite{li2021ai}.
First, we analyze the choice of $\hat{u}$ (\refFig{user_study} right) and $\hat{t}$ (\refFig{diffusion_timestep} left) for extraction of $\featsc$ using PA-MPJPE.
We observe the best performance for $\hat{u} = 3$, which we consequently use for both VDMs in our other experiments. 
We further find our method is not sensitive to the choice of $\hat{t}$ above $t \approx 15$.
Therefore, we select $\hat{t}=20$ for VC and $\hat{t}=40$ for DC.

Next, we ablate our regularization losses (\refTbl{losses}).
As expected, the smoothness of $\mathcal{L}_s$ reduces the Acceleration error, while $\mathcal{L}_f$ reduces shape distortions recorded by the Per-Vertex Error.
In contrast, the depth regularization of $\mathcal{L}_d$ does not lead to an improvement in performance metrics, but we observe that it discourages perceptually-objectionable depth errors (\refFig{losses} right).

\highlight{Finally, in \refFig{vertex_ablation_small} we ablate the mesh resolution in our method with NJF. 
We find that the output quality degrades gracefully and predictably with reduced vertex count which is favorable for our method's versatility. 
See \refSupp{number_of_vertices} for an extended discussion.}

\section{Discussion}
\label{sec:discussion}

\begin{figure}
    \includegraphics[width=0.95\linewidth]{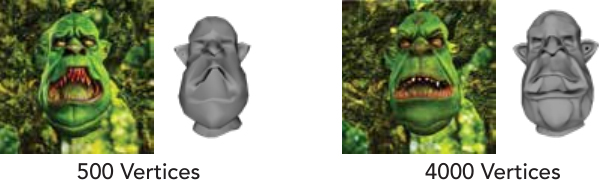}
    \caption{Effect of number of vertices on our method. Model source: Jaka Ardian 3D art / model from Indonesia.}
    \label{fig:vertex_ablation_small}
    \hspace{5mm}
    
    \centering
    \includegraphics[width=0.95\linewidth]{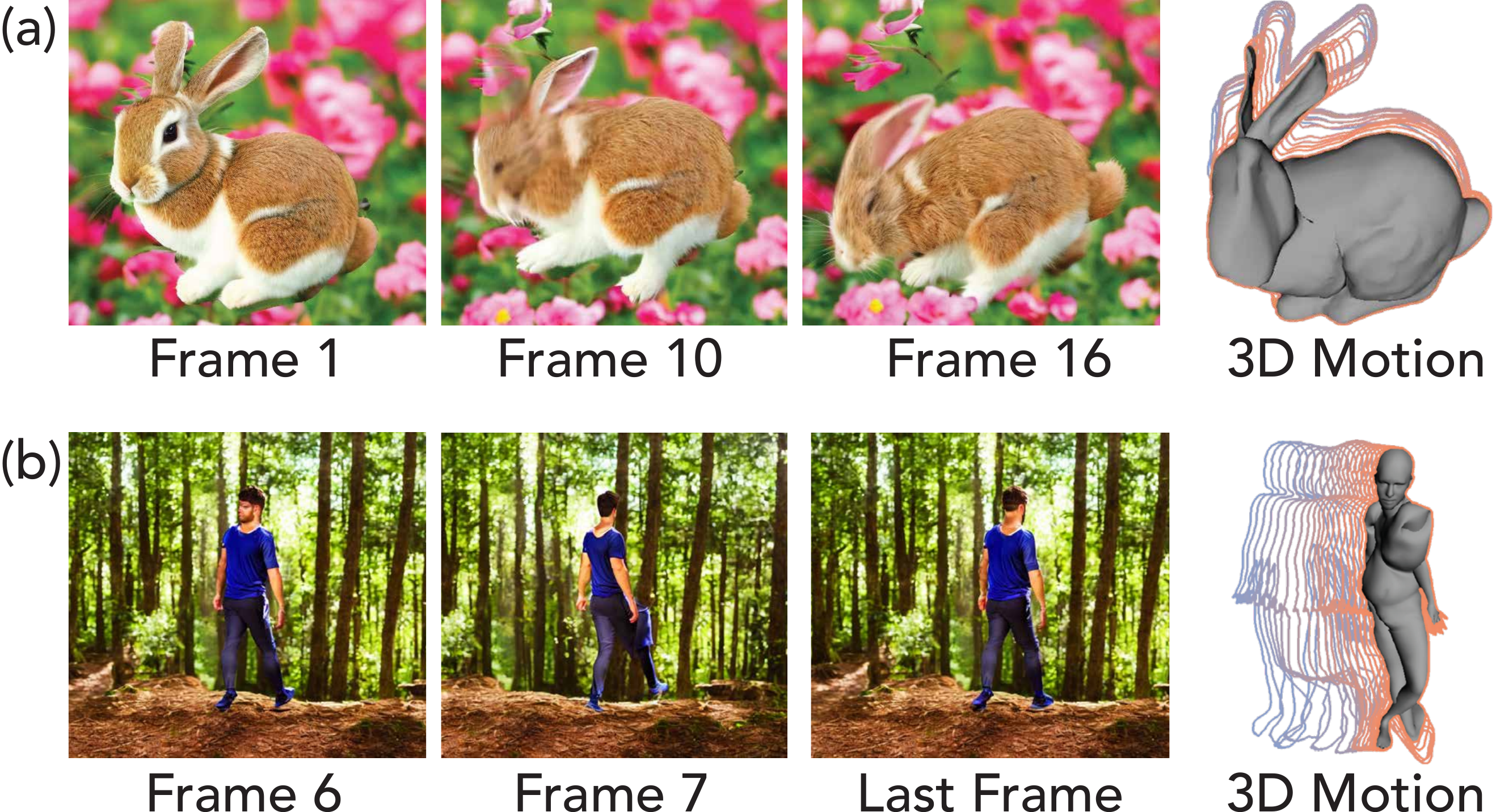}
    \caption{Failure cases showing frames of the VC VDM output and our fitted motion. 
    (a) The VDM produces fast motion accompanied by ear disappearance that our model explains as an undesired head deformation.
    (b) The VDM suddenly flips body orientation by 180 degrees which confuses our tracking and leads to self-intersections.}
    \label{fig:failure_cases}
\end{figure}

\paragraph{Limitations and Future Work}
\label{sec:limitations}
Single-view motion supervision struggles to resolve motion-in-depth or occlusions, which we mitigate using regularization at a risk of overall motion reduction (see \refFig{losses} right). 
\highlight{We acknowledge this as a limitation and a motivation for further research which could offer an improvement through multi-view supervision at the cost of additional training data~\cite{kapon2023mas}.}
%Severe occlusions can break algorithm, do something like Multi-View Anscestral Sampling. 
We demonstrate a zero-shot method supporting a range of animation models, but we acknowledge that the high degree-of-freedom in NJF permits undesired distortions (see \refFig{failure_cases}a).
These could be potentially remedied though a static shape supervision inspired by 3D generative models~\cite{lee2024text} with a possible diversity reduction stemming from SDS~\cite{liang2023luciddreamer}.
On top of this, the motion produced by the current VDMs might not adhere to the prompt or might contain physically impossible transitions (see \refFig{failure_cases}b). 
To counter this, the fast run-time of our method could be combined with a suitable rejection heuristic. 
Furthermore, we expect to benefit from future VDM improvements~\cite{videoworldsimulators2024,googledeepmind2024veo}.
This will also allow for generating longer sequences, necessitating memory off-loading, which is currently absent in our implementation.
Finally, an interesting future direction is to close the loop and constrain the VDM generation with our simultaneously optimized 3D animation model in order to prevent any spatiotemporal distortions from emerging.

\paragraph{Conclusion}
We presented a novel generative method for zero-shot 3D animation. 
\highlight{Despite its limitations stemming from the single-view supervision, we demonstrated that it produces visually preferable motions across diverse unseen 3D shapes at computation cost lower than end-to-end 4D generative methods.}
We see our method as a capable tool for analysis of motion spaces in VDMs, and for affordable re-animation of otherwise static 3D assets in large-scale virtual environments.

\paragraph{Ethical Considerations}
Our method produces novel poses for 3D objects including human bodies and faces, 
but we do not focus on realistic appearance modeling.
The biases in backbone VDMs can influence our method and are a priority research interest to the community.

{\small
\bibliographystyle{unsrtnat}
\bibliography{bibliography}
}

\appendix
\newpage

\clearpage
\appendix

\twocolumn[
\centering

\section*{\LARGE Appendix \vspace{0.1cm} \\ \Large  MotionDreamer: Exploring Semantic Video Diffusion features for Zero-Shot 3D
Mesh Animation}
\vspace{0.5cm}
\large
Lukas Uzolas \hspace{0.2cm} Elmar Eisemann \hspace{0.2cm} Petr Kellnhofer\\
\vspace{0cm}
Delft University of Technology
\vspace{0cm}\\
The Netherlands
\vspace{1cm}
]

\section{Additional Implementation Details}
\label{supp:impl_details}

We optimize the pose fitting for $1\,000$ iterations.
We initially optimize only $\defp^0$ and linearly increase the number of optimized frames from $1$ to $L$ between iterations $0$ and $500$.
We use a constant learning rate of 0.0005 and Adam optimizer~\cite{kingma2014adam} in Pytorch~\cite{paszke2017automatic}. 
Our MLP $m$ consists of $6$ layers, each with a hidden dimension size of $256$, and we scale the final output by a constant $\alpha=0.01$. 
We apply a frequency encoding~\cite{mildenhall2020nerf} for the input $l$: $\gamma(.) = (l, sin(2^0\pi l), cos(2^0\pi l), ...,  sin(2^{k-1}\pi l), cos(2^{k-1}\pi l))$ with $k=6$. 

For each input shape $\mesh$, we define the canonical camera $\cam$ manually.

\subsection{Video Diffusion models}
\label{supp:vdms}

We use the official implementations for VideoComposer~\cite{wang2024videocomposer} (VC), DynamiCrafter~\cite{xing2023dynamicrafter} (DC), as well as for Stable Video Diffusion~\cite{blattmann2023stable} (SVD), where the latter two are accessed through the Diffusers library~\cite{von-platen-etal-2022-diffusers}.
We adopt the same hyperparameters for all wherever possible.
We set the classifier-free guidance~\cite{ho2021classifier} to 6 and we generate 16 frames with an assumed framerate of 16 fps.
We use the recommended schedulers with $T=50$ inference steps.
We discuss the choice of conditioning images for each model and the omission of SVD from our experiments in \refSupp{texturing_effect}.
Finally, we observe that VC provides faster inference than DC, which is why we adopt it for our quantitative experiments that require large number of optimizations.

\paragraph{Matching rendered image to VDM features for DC}
\label{supp:correspondence_finding_dc}
Unlike VC and SVD, Dynamicrafter does not enforce the input image to be the first frame of the output video, which is an assumption of our method.
This is because all VDM frames are initialized with the same embedded input image:  $\mathcal{E}(\mathbf{x}) = \mathbf{z}^0 = \mathbf{z}^1 = ... = \mathbf{z}^L$ and then they drift during the inference. 
We observe that this drift is minimized for the output frame matching the input image and hence we explicitly detect the frame $l^*$ where features change the least between the inference steps $t$:
\begin{equation}
    l^* = \mathop{\mathrm{arg\,max}}_l \sum_t \frac{\kappa(\mathbf{\featsc}^l_t, \mathbf{\featsc}^l_{t-1}) - \mu_{\mathbf{A}_{\kappa, t}}}{\sigma_{\mathbf{\featsc}_{\kappa, t}}},
\end{equation}
where $\mu_{\mathbf{\featsc}_{\kappa, t}}$ and $\sigma_{\mathbf{\featsc}_{\kappa, t}}$ are the mean and standard deviation of the cosine similarities of activations $\mathbf{\featsc}_t$ at step $t$. Finally, $l^*$ can be used as the frame index for feature reprojection.

\subsection{Animation Models}
\label{supp:anim_models}
We experiment with four different animation models.

\paragraph{SMPL} 
Skinned Multi-Person Linear~\cite{SMPL:2015} is a skinned mesh-based human model that supports various body shapes and human poses. Vertices are deformed based on forward kinematics and linear blend skinning: $\mathbf{u}_i^l = \sum_b w_{b,i}\mathbf{T}^l_b\mathbf{u}^\textrm{init}_i$, where $\mathbf{T}^l_b \in \reals^{4 \times 4}$ is the roto-translation of bone $b$ at time step $l$ and $w_{b,i}$ the skinning weight determining how strongly vertex $\mathbf{u}_i$ is attached to $b$. $\mathbf{T}^l_b$ is defined recursively by its parent bone transformation according to a kinematic hierarchy.

\paragraph{SMAL} 
SMAL~\cite{zuffi20173d} is another skinned model that can represent various quadrupedal animals, namely lions, cats, dogs, horses, cows and hippos. It follows the sample approach of forward kinematic and linear blend skinning for reposing as SMPL. We make use of the SMALify \cite{biggs2020left} implementation in our work.

\paragraph{FLAME} 
FLAME~\cite{FLAME:SiggraphAsia2017} also adopts the SMPL formulation but expands it by articulation of the jaw, and the eyes. 
It utilizes blend-shapes to model facial expression offsets for all vertices in the mesh: $\mathcal{U}_{exp} = \sum_n^{\vec{|\psi|}} \vec{\psi}_n \mathbf{E}_n$, where $\vec{\psi}_n$ denotes the n'th expression coefficient,  $\mathcal{E} = [\mathbf{E}_n,..., \mathbf{E}_{\vec{|\psi|}}] \in \reals^{3N\times \vec{|\psi|}}$ is the orthonormal expression basis, and $\mathcal{U}_{exp}$ contains the vertex expression offset for each $\mathbf{u}_n$. We further find it beneficial to scale the expression coefficient $\vec{\psi}_n$ by a factor of $5$ in FLAME

Note that we keep the shape parameters fixed for SMPL, SMAL, and FLAME. Please refer to the corresponding work for more details.

\paragraph{Neural Jacobian Fields (NJF)} 
Our method also supports arbitrary meshes that are neither rigged nor have blendshapes. To animate these types of meshes we make use of NJF~\cite{aigerman2022njf}. In NJF, the deformation is obtained by indirectly optimizing the per-triangle Jacobians $J_i \in \mathbb{R}^{3\times 3}$ for each face $f_i$, instead of directly regressing the displacement for each vertex. To retrieve the deformation map $\Phi*$, a Poisson problem is solved: $\Phi* = \min_{\Phi} \sum_{f_i}  |f_i|\parallel \nabla_i(\Phi) - J_i \parallel_2^2,$ where $\nabla_i(\Phi)$ is the Jacobian of $\Phi$ at triangle $f_i$ and $|f_i|$ represents the area of the triangle. 
We follow the implementation of \citet{gao2023textdeformer}, and initialize the Jacobians with identity matrices. Besides the Jacobians, we additionally optimize root rotation, center of rotation, and a global translation vector. We also makes use of the Jacobian regularization \cite{gao2023textdeformer} to avoid diverging too far from the initial geometry.
Consequently, we expand our full optimization objective with an additional term
$$
\mathcal{L}_{j} = 1/(2M) \sum_i (\parallel J_i - I \parallel_2) + \parallel J_i - I \parallel_1),
$$ 
where $M$ is the number of triangle faces.
Therefore, for NJF, we minimize $\mathcal{L^\prime} = \mathcal{L} + w_{j} \mathcal{L}_{j}$, where $w_j = 0.5$.

Our requirements for the inputs mesh are entirely dependent on the animation model.
For NJF, we assume a mesh with a single connected component.
For multi-component meshes, we adopt the preprocessing from \citet{Wang-2022-dualocnn} and transform the mesh representation into an SDF and resample the mesh based from this SDF. 
We additionally decimate faces through Quadric edge collapse~\cite{garland1997surface} to reach $8\,000$ vertices.
In practice, we observe that this procedure is robust even for meshes that are not perfectly watertight nor 2-manifold.

For all animation models, we scale the global translation vector $\mathbf{t}$ by $0.1$.

\section{User Study}
\label{supp:user_study_details}

\subsection{Baseline methods}
\label{supp:baselines}

\paragraph{DG4D}
We use the original implementation provided by \citet{ren2023dreamgaussian4d} but adapt two hyperparameters such that the model can be trained with only 24\,GB of VRAM. Namely, we reduce the batch size from 14 to 8 and and the number of views per step (\textit{n\_views}) from 4 to 2. 
In its original setup, DG4D automatically removes the background of the input image before passing it to a VDM with a tool \emph{Rembg}\footnote{https://github.com/danielgatis/rembg}. However, as we show in \refSupp{texturing_effect}, VC produces better results for input images with background. 
Therefore, for VC, we remove the background after the video generation instead by applying Rembg to each video frame.
When using DC, the pipeline of DG4D is unaffected.

\paragraph{MDM-MT}
We observe a lack of class-agnostic end-to-end pure motion generators.
Therefore, we combine a human-specific motion generator with a general motion transfer method while accepting that the performance of such solution will depend on morphological and semantic proximity of the source and target shape class.
To this goal, we first use a pre-trained author's implementation of the text-conditioned motion diffusion by \citet{tevet2023human} to generate a unique 2D skeletal human motion sequence for each example in our study.
We adapt the motion text prompts used for our method (\refTbl{vmd_prompts}) to the human domain using a template ``a person is [ACTION]'' e.g., ``a horse is walking'' $\to$ ``a person is walking''.
Next, we use a 2D-to-3D human body pose uplifting method adapted from the code of \citet{zuo2021sparsefusion} to obtain sequence of SMPL~\cite{SMPL:2015} meshes.
Finally, we follow the procedure and code of \citet{liao2022skeleton} to retarget the SMPL animations to our target meshes.
We aply this step consistently even for the SMPL target mesh.
Finally, we render the first 16 frames of the resulting mesh sequences in the same way as for our own method.

\subsection{Stimuli} 
\label{supp:stimuli}

In our study we utilize 10 different shape-prompt pairs (2 SMAL, 2 FLAME, 2 SMPL, and 4 Neural Jacobian Field combinations) and combine them each with 2 different VDMs resulting in 20 unique videos for each evaluated method.
We compare pairwise to 2 methods (DG4D and MDM-MT) for the first 3 questions, and we similarly compare to 2 methods (DG4D and VDM) for the last additional question. In total this produces $3\times 2 \times 20 + 1 \times 2 \times 20 = 160$ study trials.

\paragraph{Meshes}

We extract the surface models for SMPL~\cite{SMPL:2015}, FLAME~\cite{zuffi20173d} and SMAL~\cite{FLAME:SiggraphAsia2017} from their official implementations.
For SMPL, we opt to lower the arms to 45 degrees from the original T-pose, while we use the default ``zero'' pose parameters for others.
For NJF~\cite{aigerman2022njf}, we use the 4 open assets listed in \refTbl{meshes}.

\begin{table*}[!h]
\centering
\footnotesize
\caption{Mesh assets used to evaluate our method with NJF.}
\begin{tabularx}{\textwidth}{llp{2cm}p{6.5cm}}
\hline
Shape & Author & License & URL \\ \hline
Bunny & Stanford & Stanford Public & \url{http://graphics.stanford.edu/data/3Dscanrep/} \\
Lego truck & \citet{mildenhall2020nerf} & MIT License & \url{https://github.com/bmild/nerf} \\
Raptor & Gatzegar & TurboSquid Standard
&  \url{https://www.turbosquid.com/3d-models/raptor-dinosaur-model-1538088} \\
Palm tree & mr\_zaza & TurboSquid Standard &  \url{https://www.turbosquid.com/3d-models/3d-tropic-palm-tree-model-2090490} \\
\end{tabularx}
\label{tbl:meshes}
\end{table*}

\paragraph{VDM Target Motion Prompts}
\refTbl{vmd_prompts} lists VDM prompts used to generate the motion sequences for the stimuli in our study.

\begin{table*}[!h]
\centering
%\small
\caption{Prompts used to generate stimuli in our study.}
\begin{tabularx}{\textwidth}{lll}
\hline
Shape & Prompt \\ \hline
SMPL & \textit{``A person jumping up''} \\
SMPL & \textit{``A person walking forward''} \\
Horse (SMAL) & \textit{``A horse walking''} \\
Horse (SMAL) & \textit{``A horse jumping''} \\
FLAME & \textit{``A person laughing''} \\
FLAME & \textit{``A person being very angry''} \\
Bunny & \textit{``A bunny shaking its ears''} \\
Lego truck & \textit{``A yellow truck moving its shovel up and down''} \\
Raptor & \textit{``A raptor jumping''} \\
Palm tree & \textit{``A palm tree swaying in the wind''} \\
\end{tabularx}
\label{tbl:vmd_prompts}
\end{table*}

\paragraph{Single-View Texturing Prompts}
\refTbl{prompt_details} shows the positive and negative prompts used for Single-view Texturing as an input for the ControlNet diffusion model~\cite{zhang2023adding} for each shape in our experiments and the prompt for the Stable Diffusion XL~\cite{podell2023sdxl} background inpainting.

\begin{table*}[!h]
\centering
\small
\caption{Prompts used for our Single-View Texturing and background inpainting.}
\begin{tabularx}{\textwidth}{lXX}
\hline
                 & \textbf{Prompt} & \textbf{Negative Prompt} \\ \hline
\textbf{SMPL}             &  \textit{``A photo of a clothed person wearing pants and tshirt in front of a <background>, photorealistic, 4k, DLSR'' }       &      \textit{``grey, gray, monochrome, distorted, disfigured, naked, nude''  }       \\
\hline
\textbf{FLAME}            &     \textit{``A portrait photo a face in front of a <background>, photorealistic, 4k, DLSR, bokeh''}     &     \textit{``grey, gray, monochrome, distorted, disfigured, render, teeth, hat''}            \\
\hline
\textbf{SMAL}            &    \textit{``A photo of a <animal> in front of a <background>, photorealistic, 4k, DLSR''}    &  \textit{``grey, gray, monochrome, distorted, disfigured, render''}               \\
\hline
\begin{tabular}[x]{@{}l@{}}\textbf{Others}\end{tabular}

&   \textit{``A photo of a <object> in front of a <background>, photorealistic, 4k, DLSR''}     &   \textit{``grey, gray, monochrome, distorted, disfigured, render''}   \\
\hline
\hline
 \textbf{Inpainting} & \textit{``Background image of a <background>''} & \textit{``Person, face, animal, object''}
\end{tabularx}

\label{tbl:prompt_details}
\end{table*}

\subsection{Instructions}
\label{supp:study_instructions}
\begin{figure*}[!h]
    \centering
    \includegraphics[width=0.85\textwidth]{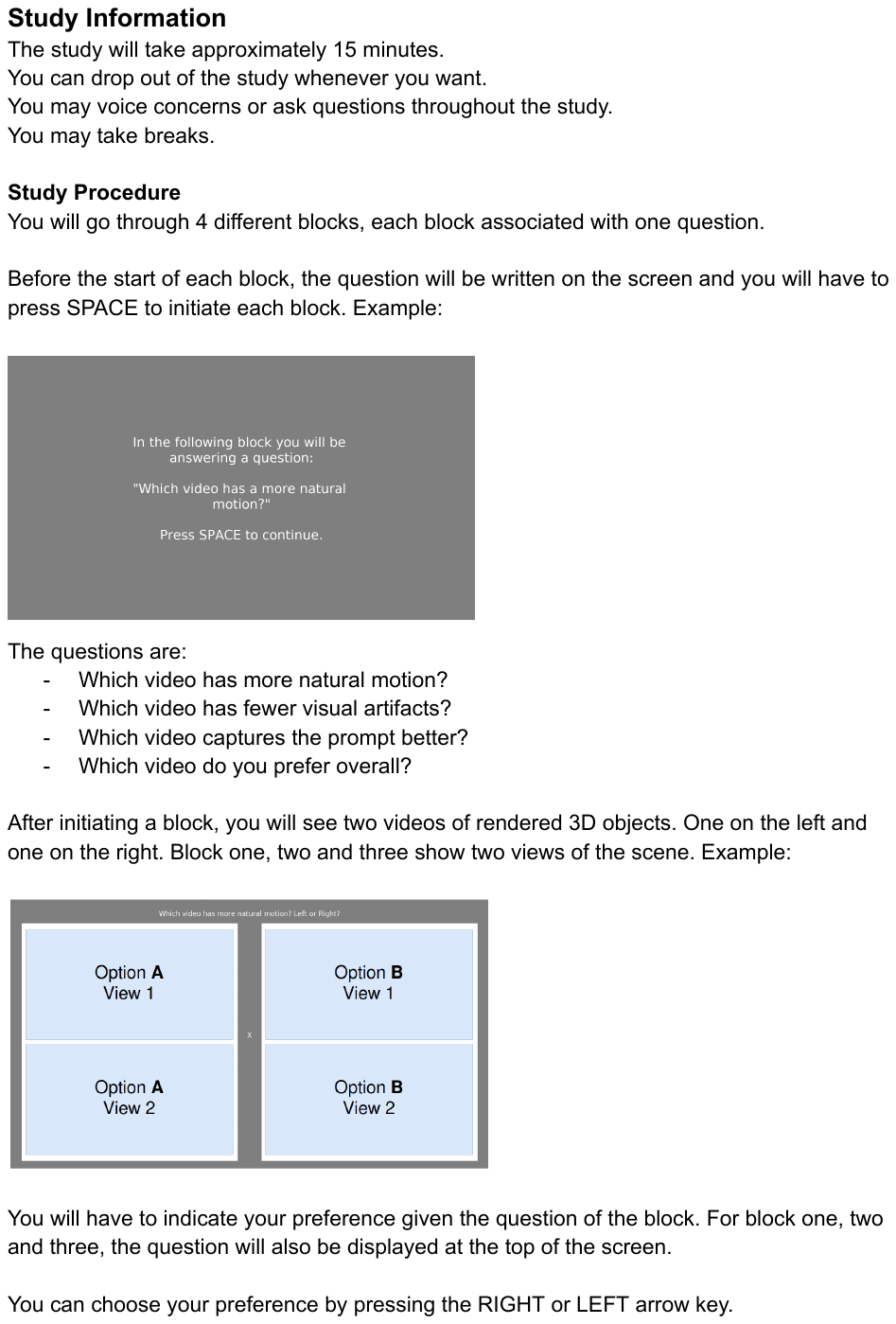}
    \caption{Study instructions that were read out and explained to our participants before the study.}
    \label{fig:study_instructions}
\end{figure*}

\begin{figure*}[!h]
    \centering
    \includegraphics[width=\textwidth]{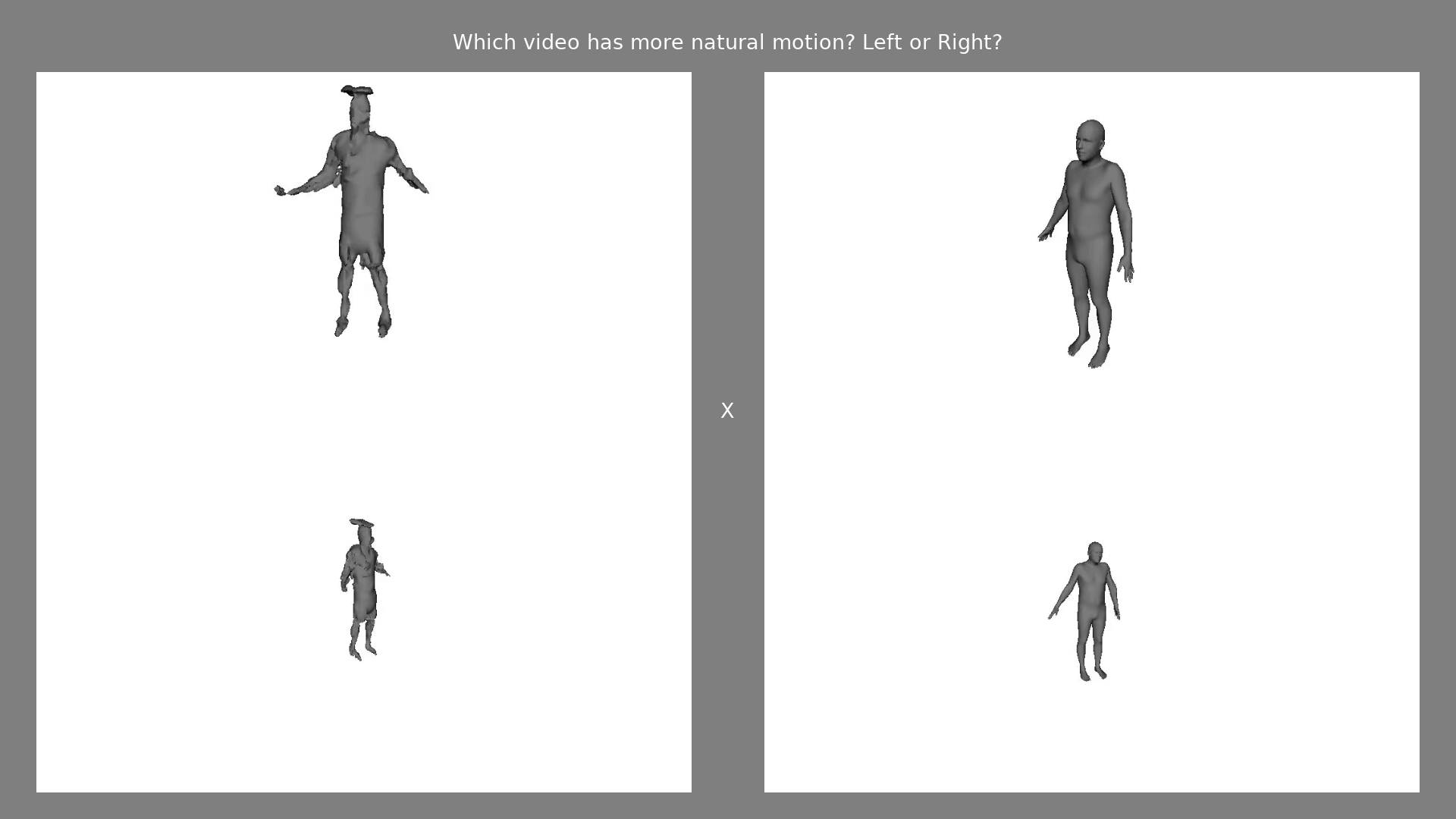}
    \caption{A screenshot of a trial for the 1st question in our user study.}
    \label{fig:question1}
\end{figure*}

\begin{figure*}[!h]
    \centering
    \includegraphics[width=\textwidth]{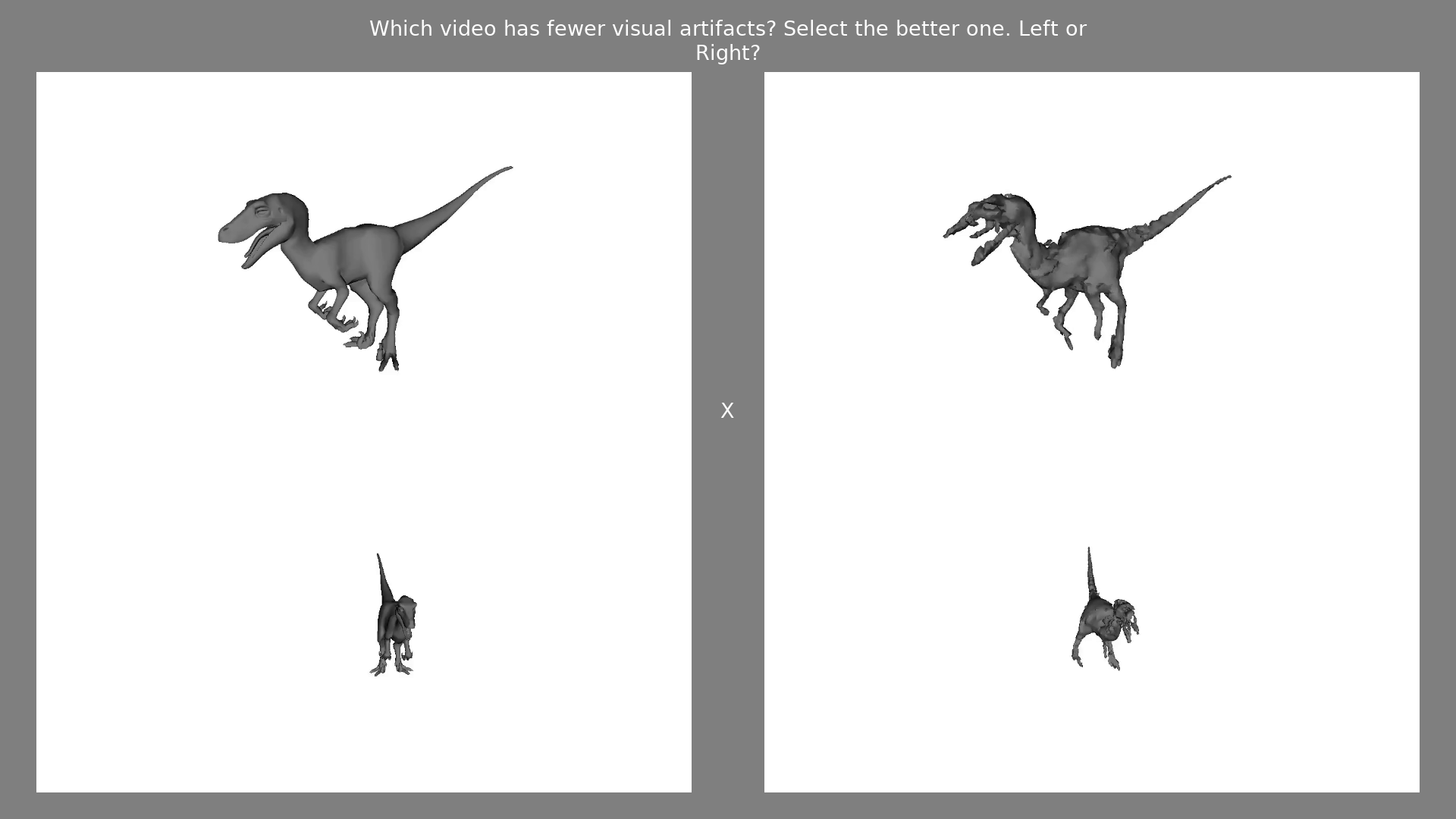}
    \caption{A screenshot of a trial for the 2nd question in our user study.}
    \label{fig:question2}
\end{figure*}

\begin{figure*}[!h]
    \centering
    \includegraphics[width=\textwidth]{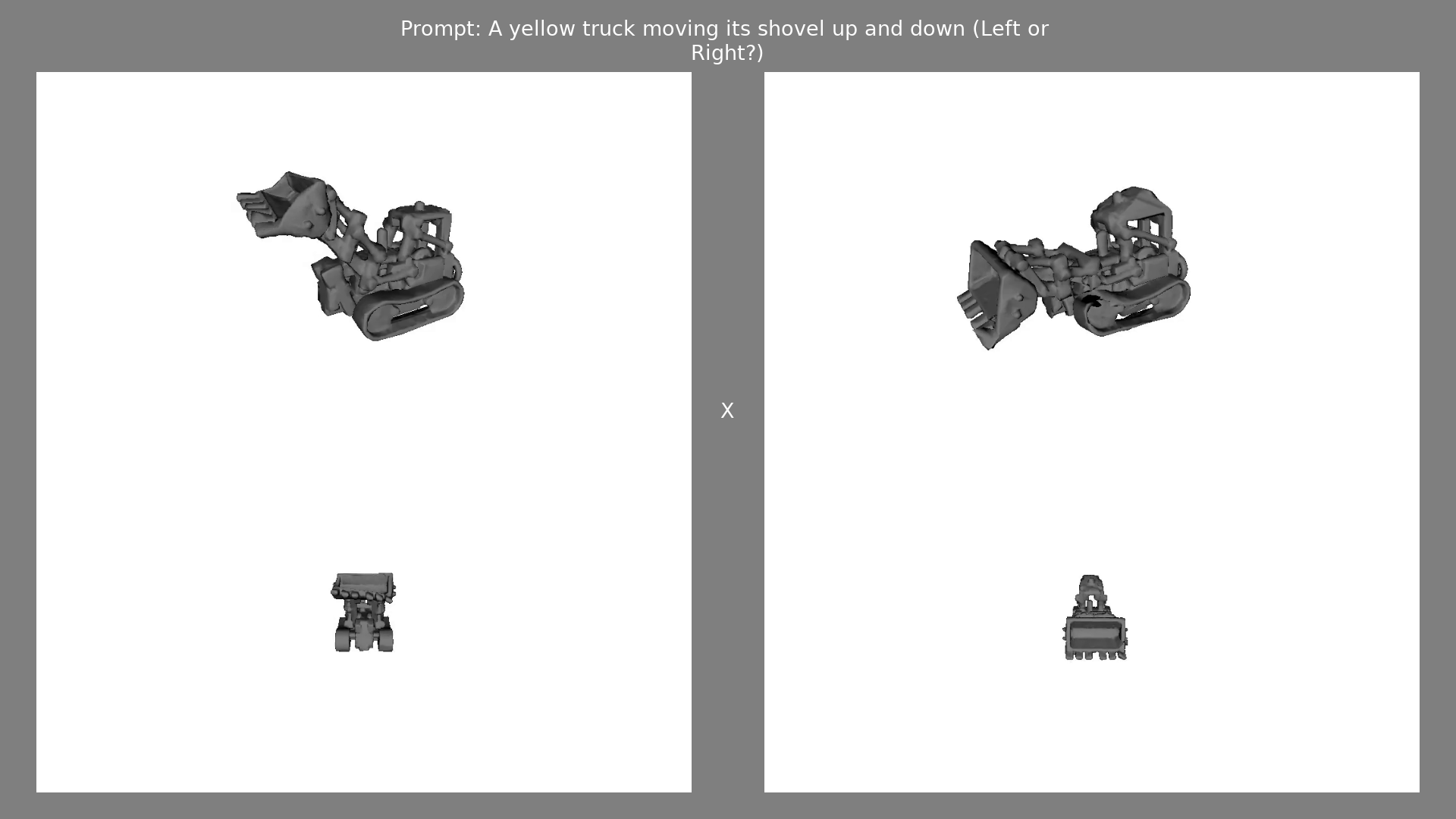}
    \caption{A screenshot of a trial for the 3rd question in our user study.}
    \label{fig:question3}
\end{figure*}

\begin{figure*}[!h]
    \centering
    \includegraphics[width=\textwidth]{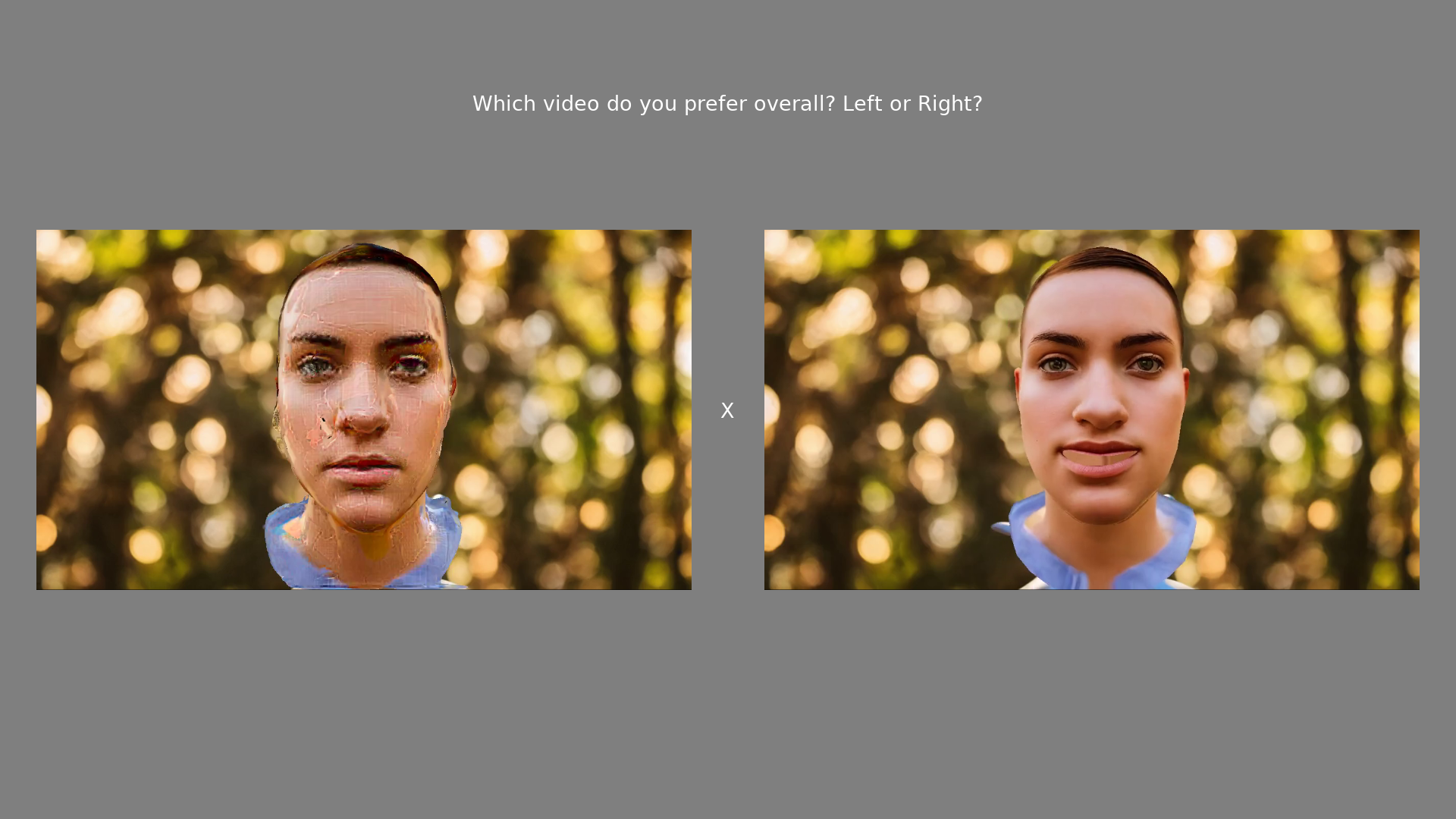}
    \caption{A screenshot of a trial for the additional 4th question in our user study.
    }
    \label{fig:question4}
\end{figure*}

\refFig{study_instructions} shows the instructions as presented to each participant before the start of the study.
\refFig{question1}, \refFig{question2}, \refFig{question3}, \refFig{question4} show screenshots of our study interface for each of the four distinct questions (3 questions in the main part and one additional question).
The questions were presented in four blocks sequentially always in the same order.
There was an instruction screen displaying the next question shown at the beginning of each block.
The order of blocks was fixed but the order and layout of the trials was randomized for each participant.

\subsection{Detailed Results}
\label{supp:study_results}

\begin{figure*}
    \centering
    \includegraphics[width=\textwidth]{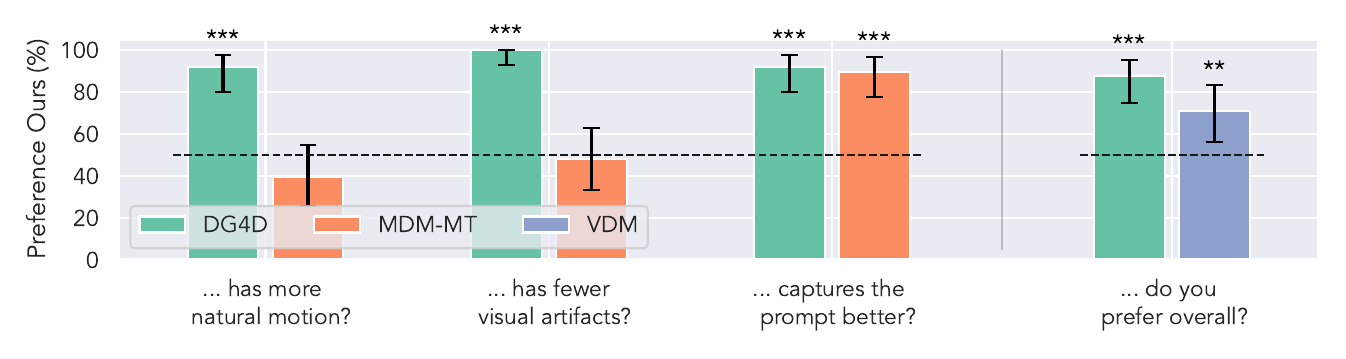}
    \caption{Study results for \textit{SMPL scenes only}.}
    \label{fig:study_results_human}

    \centering
    \includegraphics[width=\textwidth]{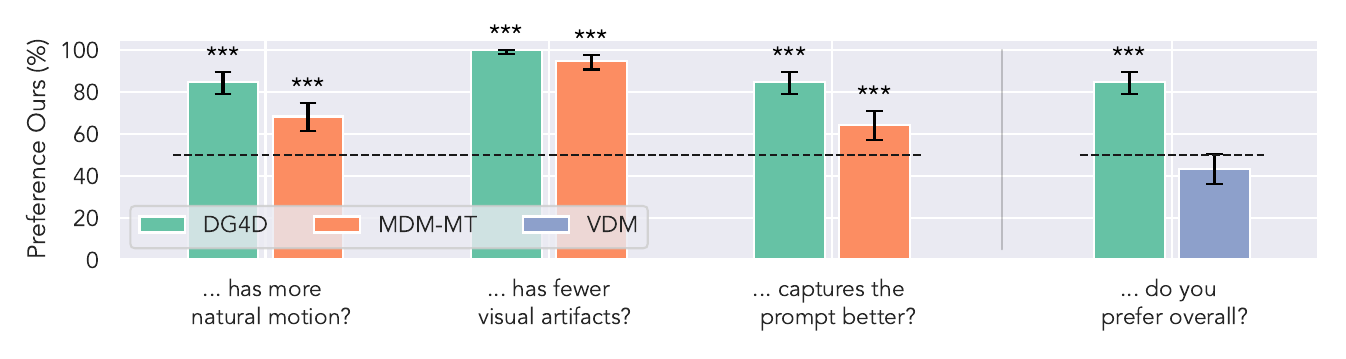}
    \caption{Study result \textit{without SMPL scenes}.
    % \todo{Recompute the study}
    }
    \label{fig:study_results_others}

    \includegraphics[width=\textwidth]{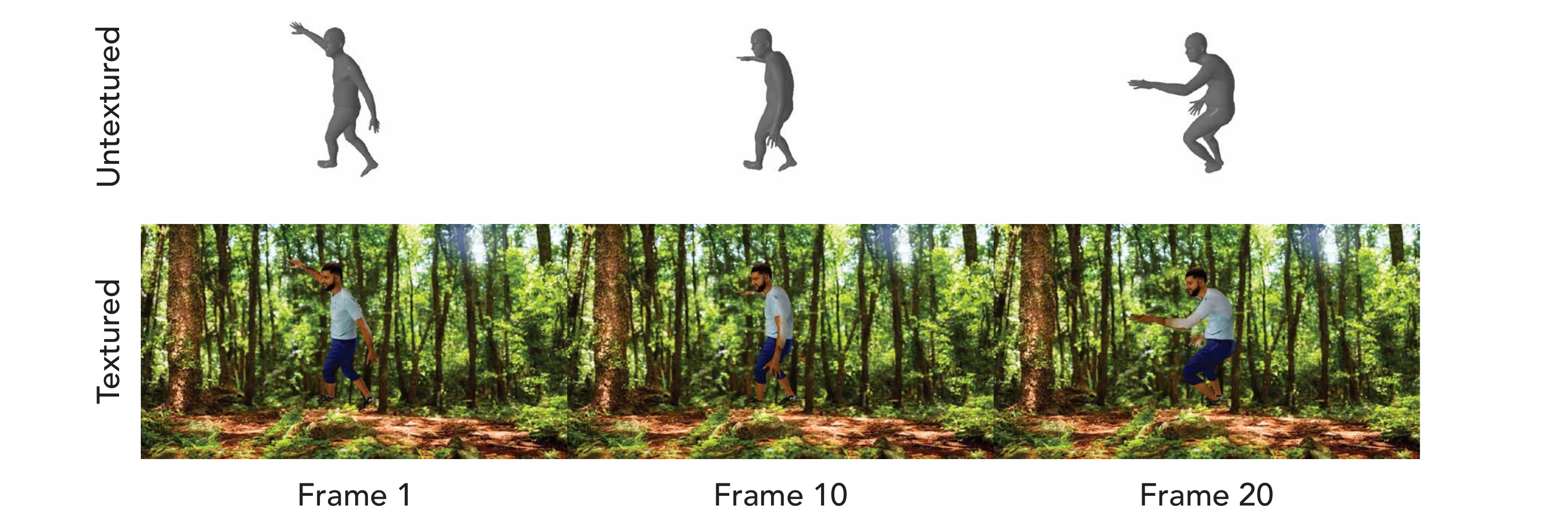}
    \caption{
    Example of rendered AIST++ scenes. On the top: The untextured models. On the bottom: Models preprocessed by our single-view texturing.
    }
    \label{fig:aist_example}
\end{figure*}

Here, we present a break-down of the results from our user study separately for the human stimuli (\refFig{study_results_human}), where the human-specific MDM-MT baseline performs well and for the remaining stimuli (\refFig{study_results_others}), where our class-agnostic method dominates.
We also offer a detailed breakdown in \refTbl{study_results_individual}.

\begin{table*}[]
\small
\centering
\caption{Breakdown of our study results showing a relative preference of our method in \%. Q1: Which video has more  natural motion? Q2: Which video has fewer visual artifacts? Q3: Which video captures the  prompt better? Q4: Which video do you prefer overall?
}
\begin{tabularx}{0.75\textwidth}{lcccccc|cc}
\hline
          & \multicolumn{2}{c}{Q1} & \multicolumn{2}{c}{Q2} & \multicolumn{2}{c|}{Q3} & \multicolumn{2}{c}{Q4} \\ \hline
          & DG4D      & MDM-MT     & DG4D      & MDM-MT     & DG4D      & MDM-MT      & DG4D       & VDM       \\ \hline
\textbf{SMPL}      & 91.7      & 39.6       & 100.0     & 47.9       & 91.7      & 89.6        & 87.5       & 70.8      \\
\textbf{SMAL}      & 54.2      & 64.6      & 100.0     & 100.0      & 77.1      & 64.6        & 85.4       & 31.3      \\
\textbf{FLAME}     & 97.9      & 97.9       & 100.0     & 100.0      & 93.8      & 95.8        & 62.5       & 29.2      \\
\textbf{Others} & 93.8      & 55.2       & 100.0     & 89.6       & 84.8      & 47.9        & 95.8       & 56.3      \\ \hline  
\textbf{All}       & 86.25      & 62.5       & 100.0     & 85.4       & 86.3      & 69.2        & 85.4       & 48.8  
\end{tabularx}
\label{tbl:study_results_individual}

\end{table*}

\section{Pose Optimization Experiment Details}
\label{supp:pose_est_study}

\subsection{Data}
We select the first $20$ frames from $20$ randomly selected human dancing motion sequences in the AIST++ dataset~\cite{li2021ai}.
Since our goal is not to reproduce the original camera poses, we use a single fixed camera $\cam$ and position the first-frame SMPL mesh into the center of its viewport.
Then we render the rest of the SMPL sequence with a fixed camera. An example can bee seen in ~\refFig{aist_example}.

\subsection{Methodology}
We follow \citet{tang2023emergent} to extract semantic features $\featsc$ from our rendered videos.
First, we add noise corresponding to a diffusion inference step $t$ to the encoded the rendered video $\mathbf{x}$: $\mathbf{z}_t = \alpha_t \encoder(\mathbf{x}) + \sigma_t \epsilon,$ where $\encoder()$ is a latent encoder for Latent Diffusion Models~\cite{rombach2022high} or identity for RGB models.
Then, we use $\mathbf{z}_t$ as an input to the VDM denoiser $f_\theta$ and obtain $\featsc$ as the U-Net activations in the same manner as in our main method (Sec.~\ref{sec:method}).

Note that we utilize MSE loss when using RGB features for optimization, as this results in better performance compared to the cosine distance.

\subsection{WHAM baseline}
To offer a fair comparison, we evaluate four different alignment strategies for WHAM~\cite{shin2023wham}, because our method starts with the known pose $\defpinit$. Results for either strategy can be found in ~\refTbl{alignment_strategies}. In strategy \textit{align}, we find the rotation and translation to align the wham output with the ground truth: 
\begin{equation}
    \hat{R} = R^{0^T}_{wham} R^0_{gt}, \text{ and } \hat{T} = diag(t)^{0^{-1}}_{wham} diag(t)^0_{gt}. 
\label{eq:alignment}
\end{equation}
The transformations are then applied to the consecutive frames l: $\Tilde{R}^l_{wham}  = \hat{R}^l R^l_{wham} $, and $\Tilde{T}^l_{wham} = \hat{T}^l T^l_{wham}$, where $\Tilde{R}$ and $\Tilde{T}$ are the new aligned root rotation and translation.

In $copy$, we copy ground truth root rotations and translations, i.e., we set $\Tilde{R}^l_{wham} := R^l_{gt}$ and $\Tilde{T}^l_{wham} := T^l_{gt}$. In \textit{full align}, we transform not only root rotations, like in \refEq{alignment}, but every bone rotation. Lastly, in \textit{copy\&align}, we copy all root rotations and translation vectors from the ground truth and also transform the bone rotations as in \textit{full align}. Note that the WHAM prediction is in full correspondence at $l=0$ in \textit{full align} and \textit{copy\&align}. We find that \textit{copy\&align} performs the best for WHAM and, therefore, adopt this alignment strategy in \refSec{pose_est}.

\begin{table}[]
\centering
\footnotesize
\caption{Evaluating different alignment strategies for WHAM.}
\begin{tabular}{lcccc}
\hline
    & \textbf{MPJPE} & \textbf{PA-MPJPE} & \textbf{PVE}   & \textbf{Accel} \\ \hline
\multicolumn{5}{l}{\textbf{\textit{Textured} (default)}}                         \\ \hline 

WHAM$_{\textit{align}}$  & .092 ± .038 & .057 ± .015   & .125 ± .047 & 8.0 ± 9.2 \\ 
WHAM$_{\textit{copy}}$  & .092 ± .038 & .057 ± .015 & .112 ± .046  & 8.0± 9.2 \\ 
 WHAM$_{\textit{full align}}$ & \textbf{.059 ± .029} & \textbf{.042 ± .016 }  & .090 ± .039 & \textbf{7.9 ± 9.0} \\ 
WHAM$_{\textit{copy\&align}}$   & \textbf{.059 ± .029} & \textbf{.042 ± .016}   & \textbf{.075 ± .036} & \textbf{7.9 ± 9.0 }\\  \hline

\multicolumn{5}{l}{\textbf{\textit{Untextured}}}                         \\ \hline 
WHAM$_{\textit{align}}$ & .091 ± .037 & .054 ± .014   & .122 ± .043 & 7.4 ± 9.1 \\ 
WHAM$_{\textit{copy}}$  & .091 ± .037 & .054 ± .014   & .109 ± .044 & 7.4 ± 9.1 \\ 
 WHAM$_{\textit{full align}}$  & \textbf{.057 ± .028} & \textbf{.039 ± .015}   & .086 ± .038 & \textbf{7.4 ± 9.1} \\ 
WHAM$_{\textit{copy\&align}}$  & \textbf{.057 ± .028} & \textbf{.039 ± .015}   & \textbf{.070 ± .035} & \textbf{7.4 ± 9.1} \\ 

\end{tabular}
\label{tbl:alignment_strategies}
\end{table}

\section{Additional Results}

\subsection{Effect of Texturing and Background for Different VDMs}
\label{supp:texturing_effect}
\begin{figure*}
    \centering
    \includegraphics[width=0.9\textwidth]{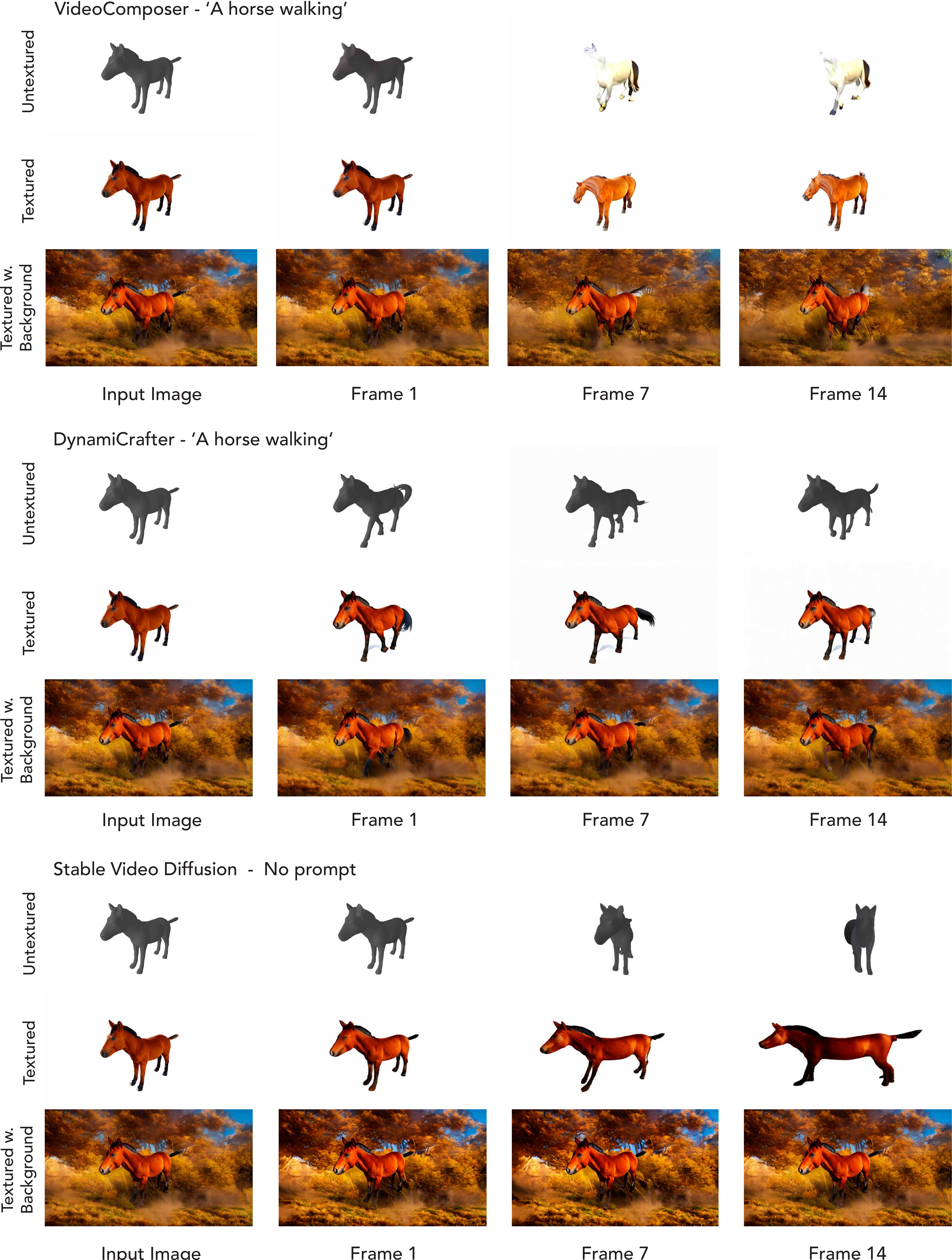}
    \caption{Comparison of 3 output video frames (columns 2--4) for 3 VDMs considered for our experiments given the same \textit{Horse} 3D mesh (1st column) rendered (from top to bottom) as an untextured shaded image, single-view textured image and a single-view textured image with a synthesized background $\bg$ (\refSec{single_view_texturing} for details of the texturing process).}
    \label{fig:horse_comparison}
\end{figure*}

\begin{figure*}
    \centering
    \includegraphics[width=0.9\textwidth]{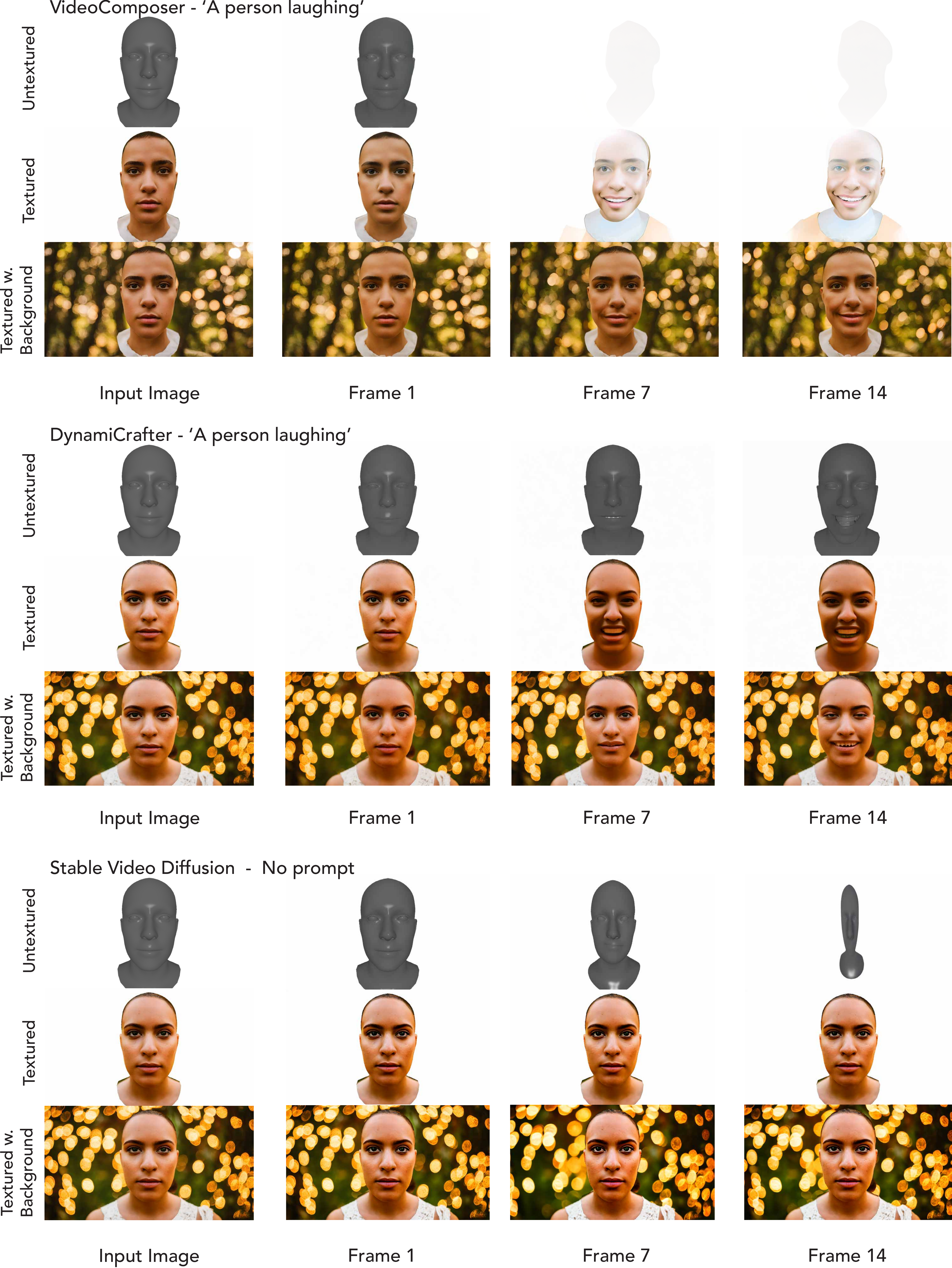}
    \caption{Comparison of 3 output video frames (columns 2--4) for 3 VDMs considered for our experiments given the same FLAME 3D mesh (1st column) rendered (from top to bottom) as an untextured shaded image, single-view textured image and a single-view textured image with a synthesized background $\bg$ (\refSec{single_view_texturing} for details of the texturing process).
    }
    \label{fig:flame_comparison}
\end{figure*}

In \refFig{horse_comparison} and ~\refFig{flame_comparison}, we compare different image input variants for the three different considered VDMs:
VideoComposer~\cite{wang2024videocomposer} (VC), DynamiCrafter~\cite{xing2023dynamicrafter} (DC), and Stable Video Diffusion~\cite{blattmann2023stable} (SVD).
We observe that VC struggles to produce coherent output for images without background images, as the object often either disappears (\refFig{horse_comparison} top) or gets distorted (\refFig{flame_comparison} top). 
DC exhibits resilience to this problem and performs well both with and without a background image.
Therefore, we opted to use images without background for DC, since it makes the videos more similar to the typical inputs of the DG4D baseline~\cite{ren2023dreamgaussian4d}.
Finally, the publicly accessible SVD model is conditioned by image only without any text prompt input. 
We observe that the motion produced by SVD for our image inputs often results in a global camera motion with no object motion. 
This is particularly prominent if no background is used.
For this reason, we excluded SVD from our other experiments.

\subsection{Qualitative Comparison to Consistent4D}
\label{supp:consistent4d}
We further compare our method to Consistent4D~\cite{jiang2023consistent4d}.
Since this is a computationally significantly more expensive method (50 minutes in its low VRAM setup versus less than 3 minutes for ours), which ened-to-end produces both the shape and the animation, we consider it a separate category from method which is better suited for quick motion prototyping and iterative animation development.
This is why we did not include Consistent4D in our user study and instead provide a general discussion and a qualitative comparison here.
A similar method DG4D was included in our study instead.

\begin{figure*}
    \centering
    \includegraphics[width=0.85\textwidth]{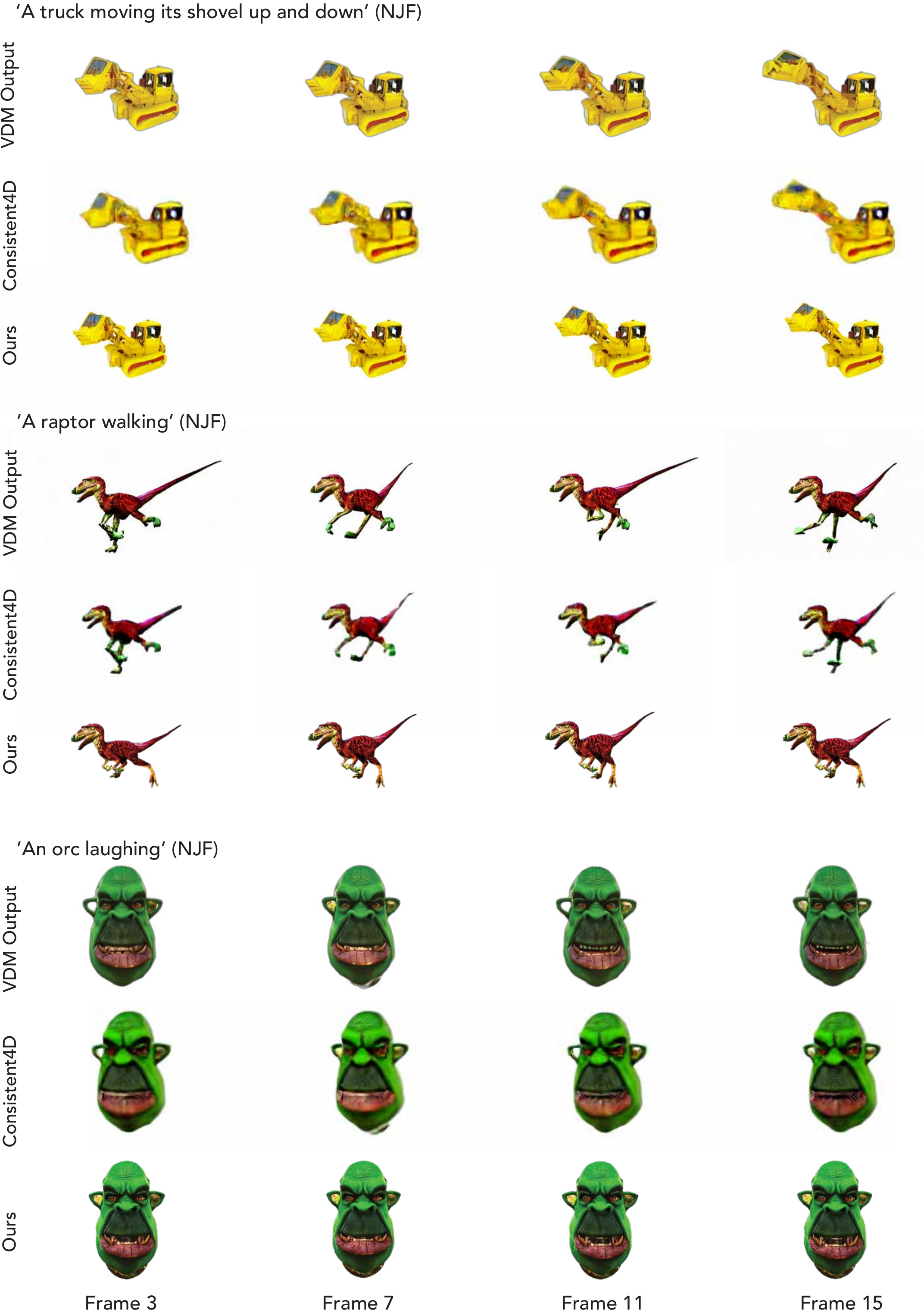}
    \caption{Comparing Consistent4D against our method.
    }
    \label{fig:consistent4d_all}
\end{figure*}

\begin{figure*}
    \centering
    \includegraphics[width=0.85\textwidth]{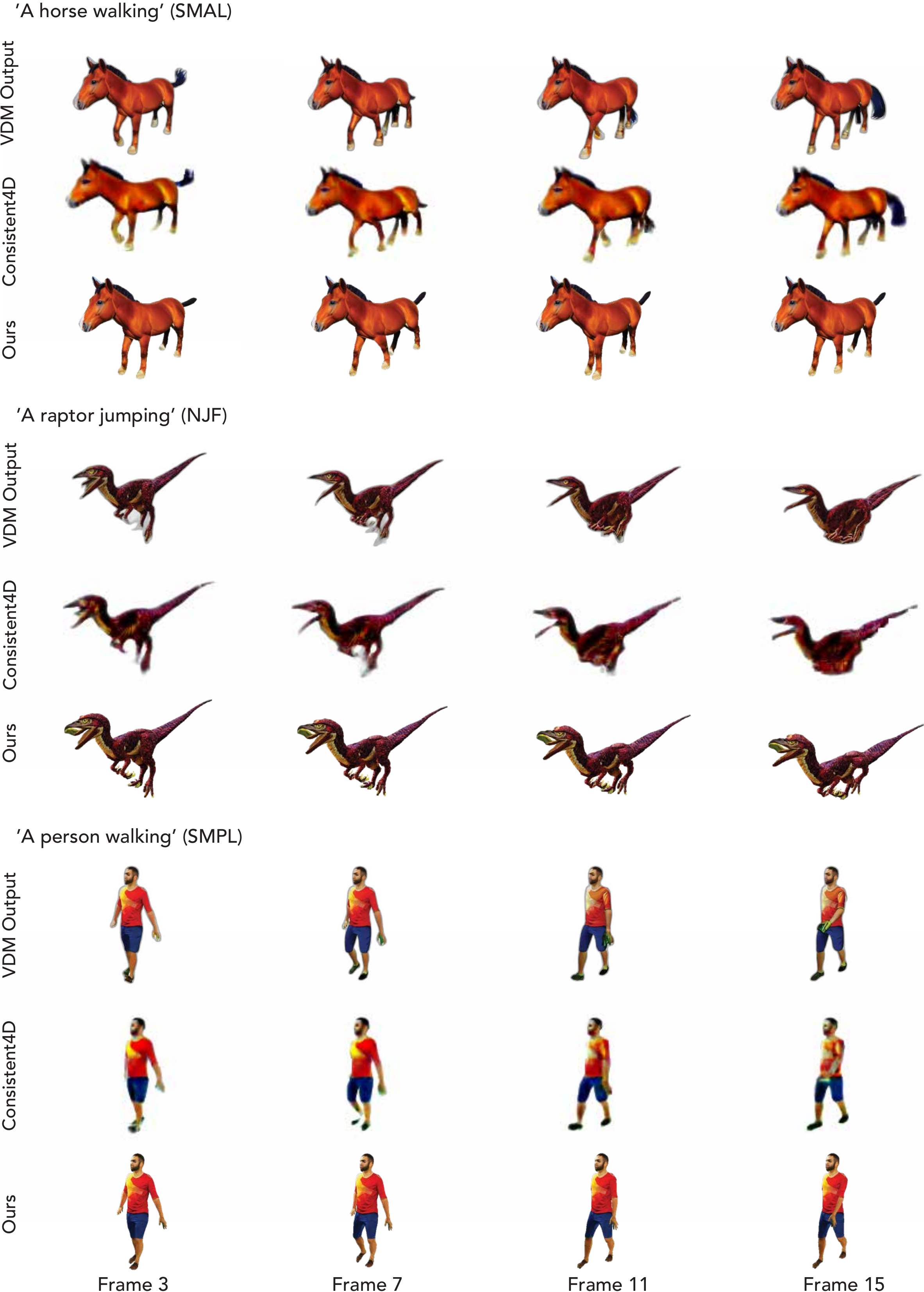}
    \caption{Comparing Consistent4D against our method.
    }
    \label{fig:consistent4d_all2}
\end{figure*}

As seen in \refFig{consistent4d_all} and \refFig{consistent4d_all2}, our method generally produces more plausible motion given the underlying geometry in a faster manner. 
The evaluation of Consistent4D take on average 32 minutes per object in its low VRAM setup, while our motion fitting takes on average under 3 minutes on an NVIDIA RTX 3090. 
Note that we exclude the time it takes to generate the driving videos in both cases. 
Consistent4D's slower runtime can be explained by its adoption of SDS which necessitates encoding the rendered RGB image into the latent space repeatedly. In contrast, our method remains in the semantic feature space.

Consistent4D utilizes a K-Plane~\cite{fridovich2023k} for their 4D representation which allows for a higher flexibility when modeling geometry distortions produced by the VDM. One such example can be seen in the raptor sequence in \refFig{consistent4d_all}. Here, the raptor's right and left legs rapidly alternate between the foreground and background, creating a sense of motion, but disrupting the raptor's underlying topology.
While the ability to fit such non-physical effects can be beneficial in some scenarios, our explicit representation is more robust to fitting to VDM artifacts and thus reduces the consequent visually implausible transformations.

Furthermore, the K-plane representation entangles shape and motion and hence it cannot be easily integrated into common computer-graphics pipelines. 
This is in contrast to our method which deforms an explicit canonical mesh, where the time-dependent vertex deformations can be easily exported. 

Lastly, Consistent4D (as well as DG4D) adopts an image-to-3D model Zero-1-to-3 \cite{liu2023zero} for 3D-Uplifting. However, Zero-1-to-3 requires input images without background which contradicts our observations that VDMs benefit from context in the background for better results (see in \refSupp{texturing_effect}).
To tackle this, Consistent4D creates masks in an automated fashion for videos with background which can, however, introduce additional errors. In comparison, our method does not rely on any such masks.

\begin{figure*}
    \centering
    \includegraphics[width=0.95\textwidth]{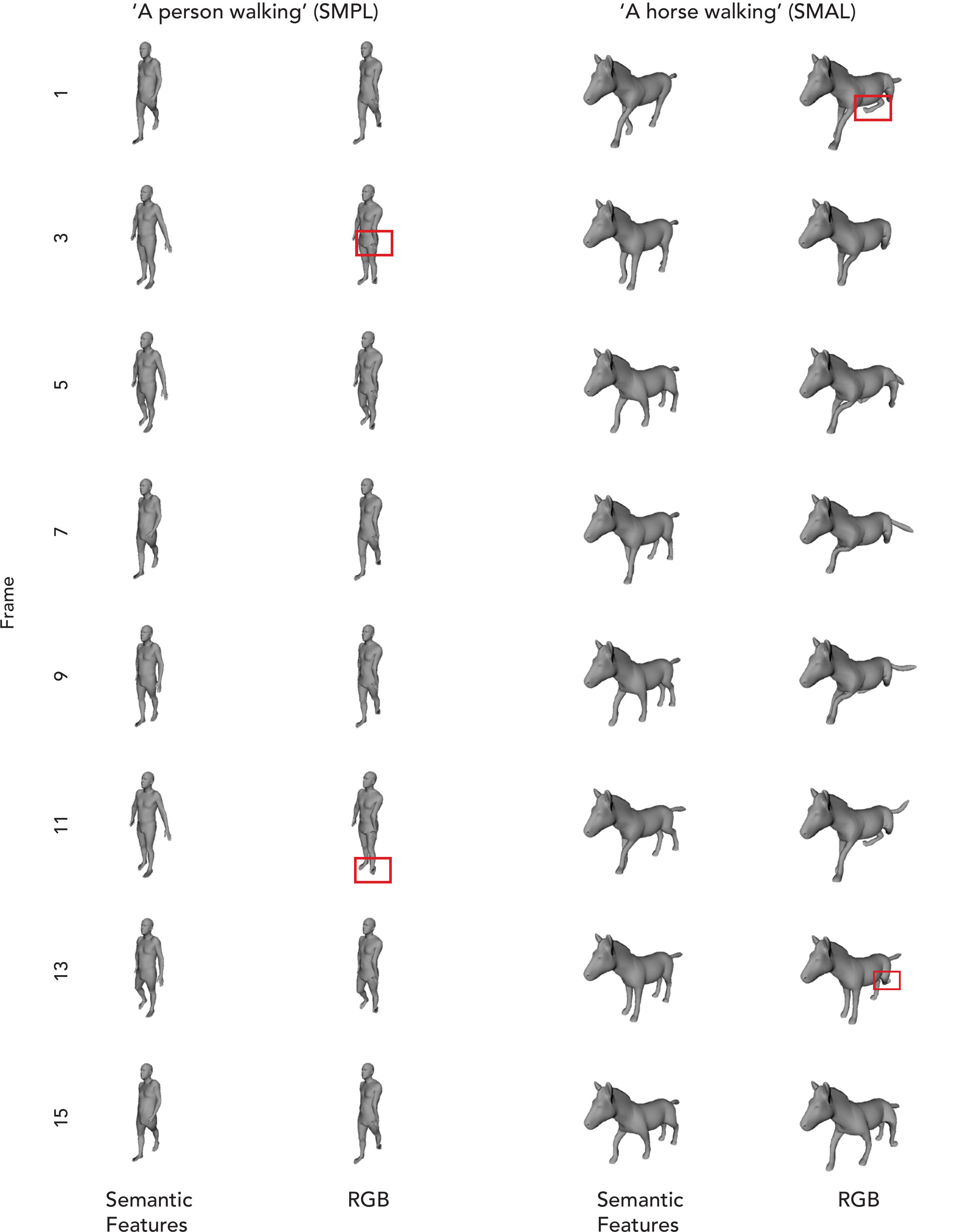}
    \caption{Comparison of semantic features against RGB used for pose optimization. Each column shows frames for a single sequence. Note the red boxes, highlighting errors in the the pose optimization when utilizing RGB: 
    1) In case of SMPL (human), the hand gets stuck in front of the torso, as the RGB features do not distinguish the body from the hand. 
    2) In case of SMAL (horse), the limbs of the horse assume less realistic articulation with RGB features.
    }
    \label{fig:rgb_vs_semantic_supp}
\end{figure*}

\subsection{Comparing Semantic Featuring against RGB for Optimization}
\label{supp:rgb_vs_semantic_features}
Here, we complement our Pose estimation experiment from the main paper and compare the RGB and semantic features end-to-end in our full pipeline
We find that modeling temporal deformations with NJF in combination with RGB features results in unstable optimization.
Therefore, we only showcase the kinematic models in \refFig{rgb_vs_semantic_supp}. 
Similarly to pose fitting, semantic features lead to superior results.

\subsection{Ablation of Number of Vertices}
\label{supp:number_of_vertices}
\begin{figure*}
    \centering
    \includegraphics[width=0.9\textwidth]{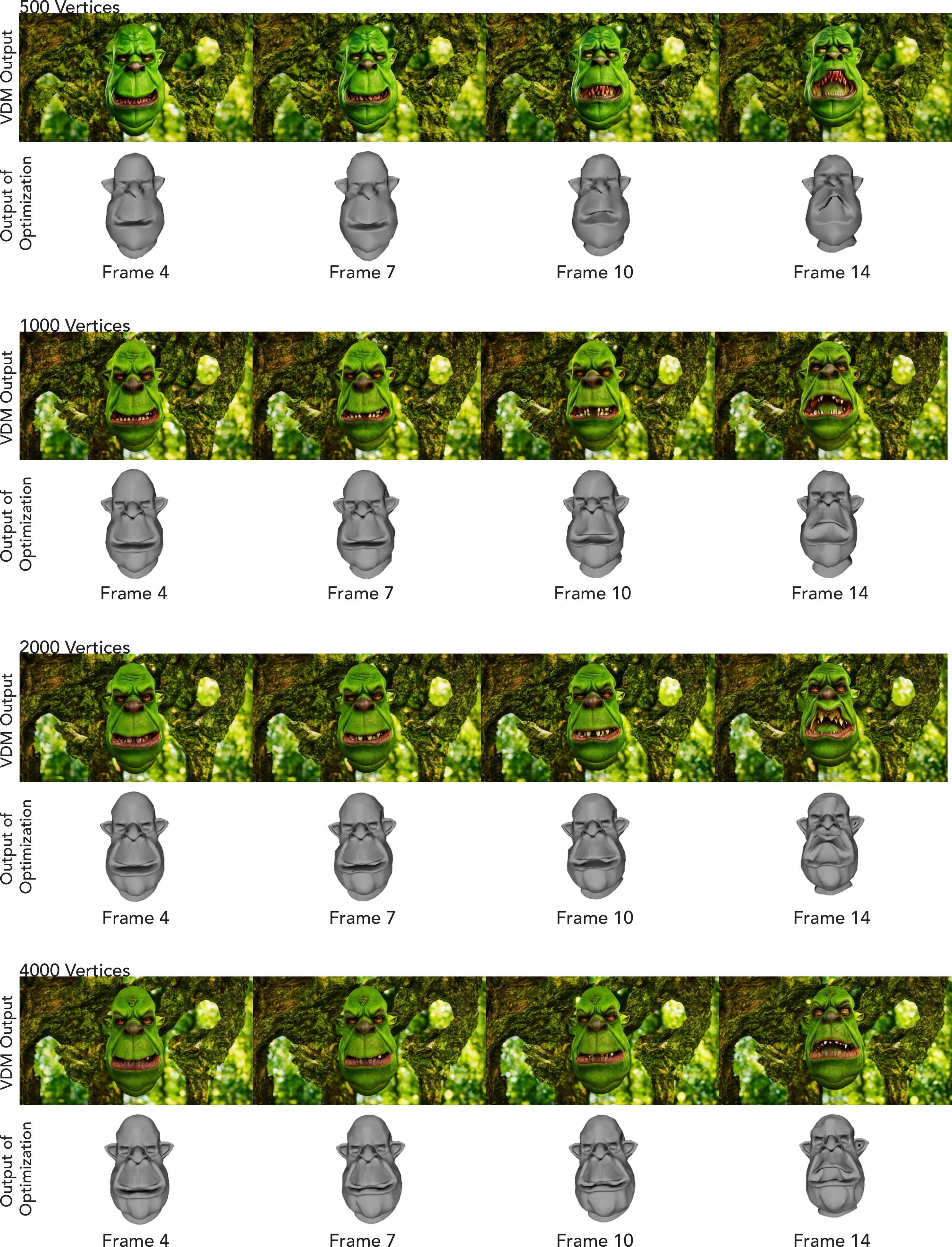}
    \caption{Effect of vertex number on our method when adopting NJF for deformations with VC as VDM backbone. Each row shows the output of our method with an increasing number of vertices. Prompt: 'An orc laughing.'}
    \label{fig:vertex_ablation_big}
\end{figure*}

\begin{figure*}
    \centering
    \includegraphics[width=0.85\textwidth]{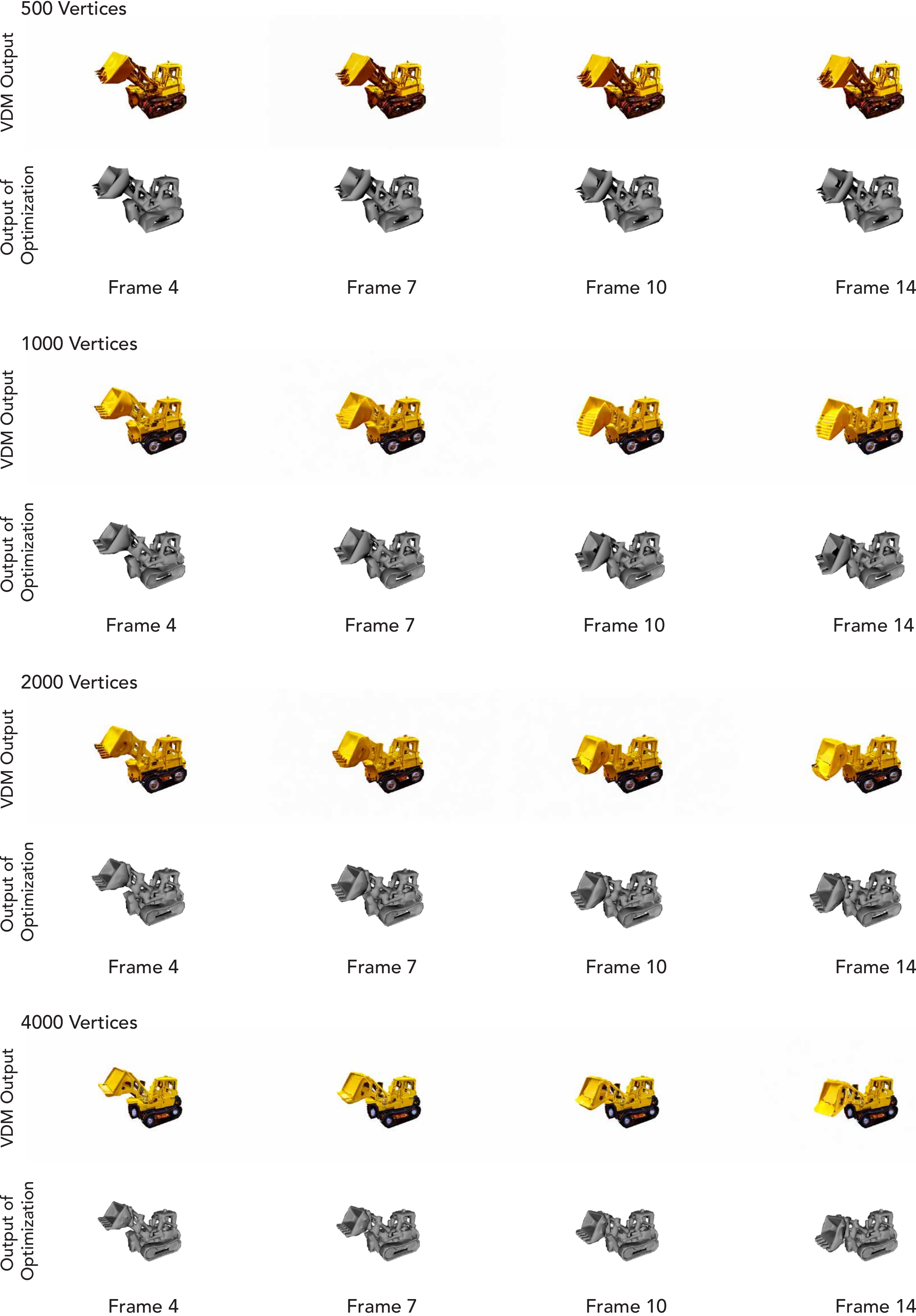}
    \caption{Effect of vertex number on our method when adopting NJF for deformations with DC as VDM backbone. Each row shows the output of our method with an increasing number of vertices. Prompt: 'A truck moving its shovel up and down.'}
    \label{fig:vertex_ablation_big2}
\end{figure*}

~\refFig{vertex_ablation_big} and ~\refFig{vertex_ablation_big2} show additional results when varying the number of vertices in our method with NJF. We find that the output quality degrades gracefully and predictably, when scaling down from 4000 to 500 vertices. 
Notice that the VDM output slightly differs, due to the variations in the conditioning input image when varying the number of vertices.

\begin{figure*}
    \centering
    \includegraphics[width=0.85\textwidth]{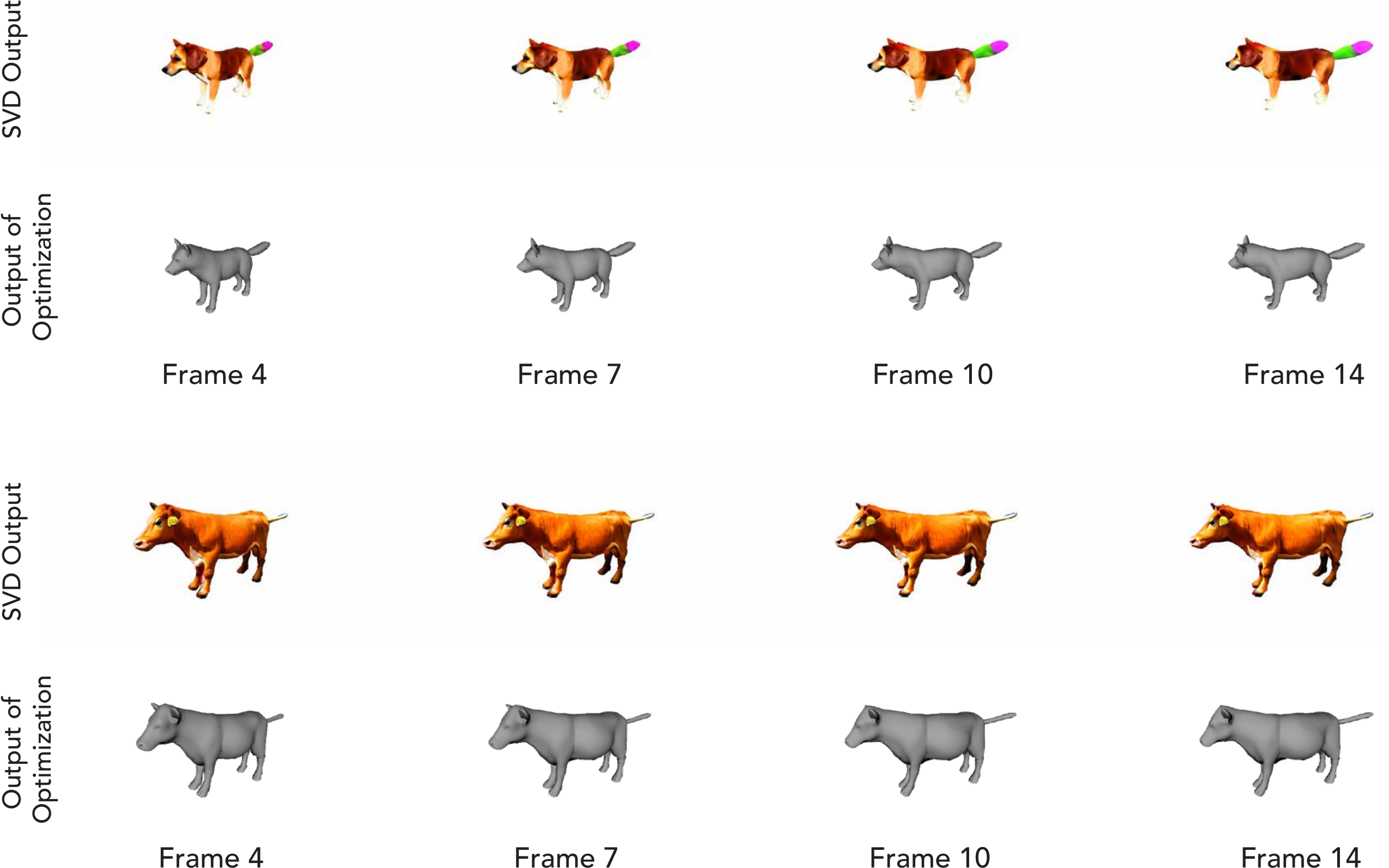}
    \caption{Results when motion fitting using the SVD semantic features. While our method is well suited to work even with SVD features, SVD videos tend to focus on a global camera rotation rather than actual object motion.
    Consequently, the fitted object motion is often minimal or uninteresting. 
    As a result we omitted SVD in our further studies in favor of other VDMs.}
    \label{fig:svd_fitting}
\end{figure*}

\subsection{Results with Stable Video Diffusion}
\label{supp:svd_results}
As reported in \refSupp{texturing_effect}, SVD produces camera motion rather than object motion. For completeness sake, we show that our method produces plausible results in \refFig{svd_fitting} when fitting to the SVD semantic features.

\end{document}